\pdfoutput=1
\documentclass[letterpaper,journal]{IEEEtran}
\usepackage{amsmath,amsfonts}
\usepackage{algorithmic}
\usepackage{algorithm}
\usepackage{array}
\usepackage[switch]{lineno}
\usepackage[caption=false,font=normalsize,labelfont=sf,textfont=sf]{subfig}
\usepackage{textcomp}
\usepackage{stfloats}
\usepackage{url}
\usepackage{verbatim}
\usepackage{graphicx}
\usepackage{multirow}
\usepackage{hyperref}
\usepackage{makecell}
\usepackage{tablefootnote} 
\usepackage{amssymb}
\usepackage{cite}
\usepackage[table]{xcolor}
\begin{document}
\setlength{\textfloatsep}{3pt plus 0.0pt minus 2.0pt}

\title{WCC: A Learning-Based Framework for Deciding When to Apply Cooperative Coevolution in Large-Scale Optimization}

\author{
Wenjie Qiu,
Hepei Li,
Jianfeng Wang,
Jun Zhang,~\IEEEmembership{Fellow,~IEEE}
and Yue-Jiao Gong,~\IEEEmembership{Senior Member,~IEEE}
\thanks{Wenjie Qiu, Hepei Li, and Yue-Jiao Gong are with the School of Computer Science and Engineering, South China University of Technology, Guangzhou 510006, China (e-mail: wukongqwj@gmail.com; csscholarpei@mail.scut.edu.cn; gongyj@scut.edu.cn).}

\thanks{Jianfeng Wang is with Taiyuan University of Technology, Taiyuan 030024, China (e-mail: wangjianfeng@tyut.edu.cn).}

\thanks{Jun Zhang is with Nankai University, Tianjin 300350, China, and also with Hanyang University ERICA, Ansan 15588, South Korea (e-mail: junzhang@ieee.org).}
}

\markboth{WCC: When to Apply Cooperative Coevolution}{Qiu et al.: WCC}

\maketitle
\begin{abstract}
Cooperative Coevolution (CC) and Non-Decomposition Algorithms (NDAs) are two primary approaches for large-scale global optimization (LSGO). Although both have achieved success, they exhibit markedly different performance across problem types and each faces approach-specific limitations, leaving  the question of ``when to use CC" unresolved. 
We address this gap with WCC, a novel learning-based framework that dynamically decides when to deploy CC in LSGO. WCC employs a streamlined design for efficient adaptive learning. 
We formulate the decision as a Markov decision process, designing a targeted feature set, a robust reward, and a tailored network architecture to provide the agent with rich information about the problem state and optimization progress. Our auto-regressive policy enables WCC to perform fine-grained, dynamic switching between CC and NDAs while conditionally allocating the computational budget without expert specifications. 
WCC shows strong competitiveness and superior generalization across problems, optimization horizons, and baseline optimizers. It realizes a ``train-once and reuse'' paradigm under a low budget, significantly curbing LSGO cost. Additionally, in-depth analyses yield insights not captured by existing theory, offering 
practical guidance on when and how to effectively deploy CC.
\end{abstract}

\begin{IEEEkeywords}
Large-scale Global Optimization, Cooperative Coevolution, Meta-Black-Box Optimization.
\end{IEEEkeywords}

\section{Introduction}

\label{sec1}
Large-scale global optimization (LSGO), typically involving over 1000 variables, represents a challenging class of black-box optimization problems with widespread real-world applications~\cite{omidvar2021review}.
The primary challenge of LSGO stems from the ``curse of dimensionality,'' where the rapidly expanding search space often renders conventional algorithms ineffective.
Cooperative Coevolution (CC) based on the ``divide-and-conquer” principle has become one of the most commonly adopted frameworks in LSGO~\cite{ma2018survey, qiu2026unicc}. 
In addition, many classical algorithms have been extended to large-scale variants to tackle LSGO, which are referred to as non-decomposition algorithms (NDAs), in contrast to those developed under the CC. 

Both CC and NDAs, despite their successes, face notable challenges. For CC, much research focuses on effective problem decomposition, where it often demonstrates competitive performance on conventional separable problems. However, CC relies on strong problem assumptions (e.g., separability and high-precision evaluations). Existing decomposition strategies have inherent biases, fail to cover all forms of separability, and are prone to decomposition errors~\cite{komarnicki2022incremental}. 
Once ideal decomposition is lost, performance drops sharply, making CC often ill-suited for real-world problems dominated by non-separability~\cite{duan2024cooperative}.
NDAs avoid complex variable decomposition and an increasing number of advanced variants have demonstrated superior performance compared to CC-based algorithms on increasingly complex problem instances. However, NDAs forgo exploiting structural priors that often exist in real problems, 
inflating the search space and computational cost, which can easily lead to optimization stagnation~\cite{loshchilov2018large}. 
As dimensionality keeps growing, a bottleneck becomes inevitable—making decomposition unavoidable~\cite{he2020mmes}.  
These differences and challenges result in highly divergent performance across problems and, in some cases, optimization stagnation in one approach—yielding outcomes that differ by several orders of magnitude.

The distinctly different optimization characteristics of both approaches have made the question of ``when to use CC'' a critical decision point. 
Many studies have theoretically analyzed the convergence of CC from perspectives such as Nash equilibrium~\cite{duan2024cooperative}. However, the black-box nature, heterogeneity, and numerous unknown characteristics of LSGO make such theoretical analysis difficult and often case-specific.
Some studies have attempted to combine these two approaches in practice based on expert knowledge. However, these hybrid methods further indicate that the question of when to use CC depends not only on the problem class but also on the optimization stage~\cite{van2004cooperative}. Applying CC at an inappropriate stage can severely degrade optimization performance, requiring analysis of phase-specific characteristics and NDA behavior, thereby compounding theoretical challenges~\cite{qiu2025novel}. 
To summarize, existing methods rely on heuristic, problem-specific switching schedules with limited cross-problem generalization, leaving the ``when to use CC'' decision an open bottleneck.

Recognizing that determining ``when to use CC'' is essentially a meta-level algorithm design problem rather than one concerning the internal mechanisms of CC, we introduce Meta-Black-Box Optimization (MetaBBO)~\cite{ma2025toward, Qiu2026Automated} into LSGO.
As an emerging research direction in recent years,
MetaBBO follows a bi-level structure, consisting of a meta-layer and an optimization-layer, and aims to automate the design of BBO algorithms by constructing intelligent agents at the meta-layer as replacements for human experts.
Many studies have demonstrated their excellent generalization ability and strong balance between exploration and exploitation \cite{song2024reinforcement, qiu2026learning}. Furthermore, the actions of their learning-based agents often provide valuable insights to researchers \cite{zhao2024automated,yi2022automated}. 
Since the decision of ``when to use CC'' is inherently not a one-shot choice, but a sequential decision-making process and a dynamic control problem under uncertainty, the appropriate decision relies heavily on the problem structure, the optimization stage and the current search status. 
To address these challenges, we formulate the process as a Markov decision process (MDP) and use deep reinforcement learning (DRL) to adaptively configure efficient algorithms for different LSGO problems. 
Instead of merely deciding whether to use CC or NDAs, the agent also determines the allocation of function evaluations (FEs) to CC and NDAs at each stage, enabling dynamic switching between the two approaches across optimization stages. To this end, we design an auto-regressive network as the agent architecture. Trained on a diverse set of LSGO problems, the agent learns to make optimal action selections at each optimization step based on a set of well-designed features. 
We refer to this architecture as the \textbf{W}hen-to-use \textbf{C}ooperative \textbf{C}oevolution (WCC) framework. The main contributions of this paper are as follows: 
\begin{itemize}
    \item \textbf{An intuitive yet effective framework to guide and analyze LSGO:} We propose WCC, an innovative framework to learn when to use CC in LSGO. Framed as an MDP and trained via DRL, our agent reduces reliance on scarce expert knowledge and offers a new perspective on this longstanding question. Unlike heuristic designs, WCC offers fine-grained decisions that explicitly determine which optimization stages should employ CC or NDAs. In addition, the framework allows complex problems to be addressed with remarkable clarity, providing the possibility of drawing insightful conclusions.
    \item \textbf{Streamlined design for efficient adaptive learning:} Based on the characteristics of LSGO, we design a robust reward function and targeted feature set for WCC to provide the agent with rich signals about both the problem and the optimization progress. An auto-regressive policy is designed to first choose CC vs. NDA, then allocate FEs conditionally, enabling step-wise switching during optimization. A network architecture is carefully designed for this purpose, enabling the agent to adaptively navigate various scenarios. 
    \item \textbf{Comprehensive generalization performance:} 
    WCC demonstrates generalization not only across different LSGO problem groups, but also across optimization horizons and baseline algorithms: WCC can be trained once under low FEs and  instantiated with different underlying optimizers, substantially reducing WCC’s training cost. 
    \item \textbf{Insightful analysis for future advancements:} 
    Thanks to its interpretable decision process, WCC further reveals clear patterns of when to deploy CC: as variable coupling increases, WCC increasingly favors NDAs, while its temporal preference shifts from early NDA to later CC on separable problems and from early CC to later NDA on non-separable ones. These observations indicate that effective CC deployment should jointly depend on problem structure and optimization stage rather than a fixed switching rule.
\end{itemize}

The remainder of this paper is organized as follows: Section~\ref{sec4} presents the methodology, Section~\ref{Experiments} reports the results and insights, and Section~\ref{CONCLUSION AND FUTURE WORK} concludes the paper.

\section{Methodology}
\label{sec4}
WCC formulates CC/NDA deployment as an MDP, where the agent observes the current optimization state $s_i^t\in\mathcal{S}$ and selects an action $a_t\in\mathcal{A}$ that jointly determines the optimization mode and computational budget. The policy $\pi_\theta$ is learned to maximize the expected discounted optimization performance over the target problem distribution $\mathcal{P}$:
\begin{equation}
    \pi^* \approx
    \underset{\pi_{\theta}\in\Pi}{\arg\max}\,
    \frac{1}{N}\sum_{i=1}^{N}\sum_{t=1}^{T}
    \gamma^{t-1}
    \operatorname{perf}
    \left(s_i^t,\pi_{\theta}(s_i^t)\mid f_i\right),
\end{equation}
where $\gamma$ is the discount factor. We optimize $\pi_\theta$ using PPO~\cite{schulman2017proximal}.
\subsection{MDP Formulation}
\subsubsection{ State}

The state $\mathcal{S}$ comprises four complementary feature groups: problem $\mathcal{S}_{pro}\in\mathbb{R}^5$, convergence $\mathcal{S}_{conv}\in\mathbb{R}^6$, resource $\mathcal{S}_{res}\in\mathbb{R}^3$, and mode $\mathcal{S}_{mode}\in\mathbb{R}^2$. Specifically, $\mathcal{S}_{pro}$ captures problem dimensionality and structural information that characterize the suitability of CC and NDAs; $\mathcal{S}_{conv}$ describes the current optimization dynamics to support stage-wise switching between their complementary search behaviors~\cite{qiu2025novel}; and $\mathcal{S}_{res}$ and $\mathcal{S}_{mode}$ characterize resource consumption and historical mode behavior, respectively, providing the contextual information needed for adaptive hybridization. 

\subsubsection{Action}
\label{Action}
The action $\mathcal{A}$ of WCC comprises two components: a mode selector $a_\text{mode}$ (choosing between NDA and CC) and a resource allocator $a_\text{res}$ (determining the FEs budget and switching timing). Assuming $a_\text{res}$ has $\kappa$ levels, the composite action space is defined as:
\begin{equation}
    \mathcal{A}:= a_{mode} \times a_{res}, a_{mode} \in \{0,1\},a_{res} \in \{0,\dots, \kappa-1\}.
\end{equation}

\subsubsection{ Reward} 
To provide scale-robust and numerically stable feedback across heterogeneous LSGO problems, we adopt a log-ratio reward:
\begin{equation}
    r_t=\ln\left(\frac{cost_{t-1}^{*}+c}{cost_t^{*}+c}\right),
    \quad
    c=10^{-8}\max\{1,cost_t^{*},cost_{t-1}^{*}\},
\end{equation}
where $cost_{t-1}^{*}$ and $cost_t^{*}$ denote the best costs at two consecutive steps. The log ratio mitigates sensitivity to objective scales and compresses large improvements, while the adaptive smoothing term $c$ prevents numerical instability when $cost_t^{*}$ approaches zero.

\subsection{Network Design}
\begin{figure*}[t]
    \centering
    \includegraphics[width=1\linewidth]{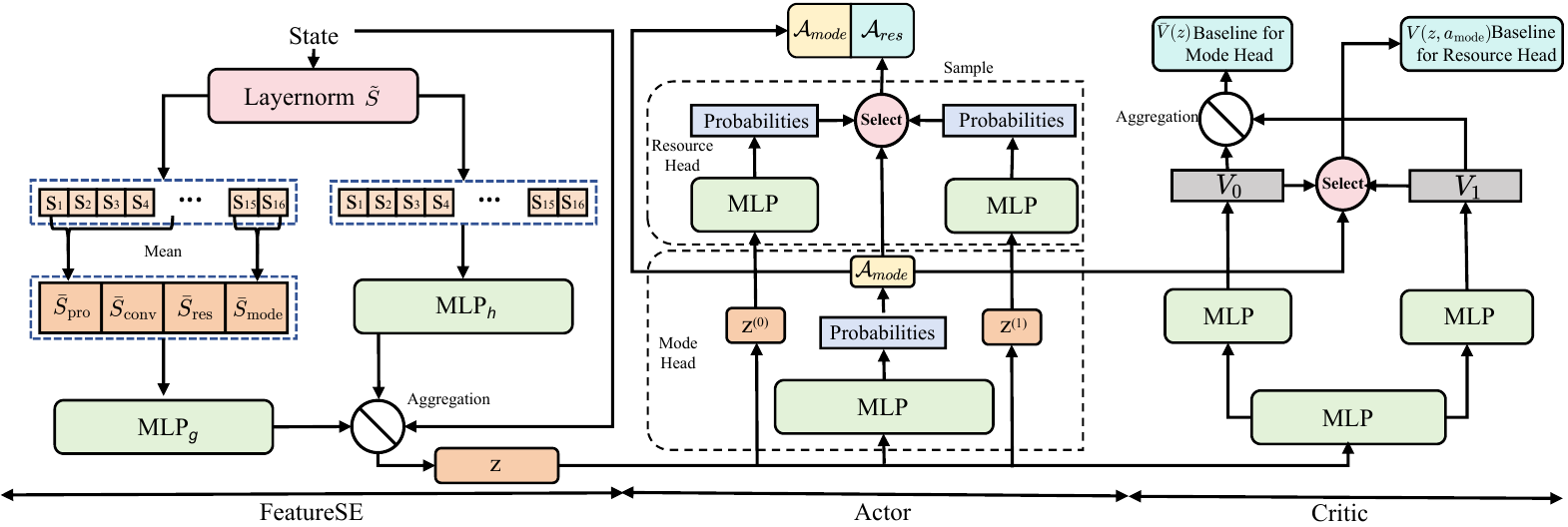}
    \caption{The network design of WCC.}
    \label{Network}
\end{figure*}

The network of WCC is composed of three components: Feature-Aware State Encoding, the Actor network, and the Critic network. 
Figure \ref{Network} shows the specific architecture.

\subsubsection{Feature-Aware State Encoding} 
Because the relevance of state information varies across problem types and optimization stages, we introduce Feature-Aware State Encoding (FeatureSE) to adaptively reweight heterogeneous state features. Given the normalized state $\tilde{\mathcal{S}}=\mathrm{LN}(\mathcal{S})$, FeatureSE learns both group-level gates $g$ over the four state categories and feature-level gates $h$ over individual features. Their effects are combined through residual gating,
\begin{equation}
    w_j=(1-\lambda)+\lambda g_{k(j)}h_j,\qquad
    z_j=s_j w_j,
\end{equation}
where $k(j)$ denotes the group of feature $j$ and $\lambda$ controls the gating strength. This hierarchical reweighting allows WCC to emphasize different information according to the current optimization context while avoiding excessive feature suppression. 

\subsubsection{ Actor}  
We adopt an autoregressive policy that first selects the optimization mode and then allocates resources conditioned on the selected mode:
\begin{equation}
    \pi(a_{\mathrm{mode}},a_{\mathrm{res}}\mid z)
    =
    \pi_{\theta_{\mathrm{mode}}}(a_{\mathrm{mode}}\mid z)
    \pi_{\theta_{\mathrm{res}}}(a_{\mathrm{res}}\mid z,a_{\mathrm{mode}}),
\end{equation}
where $z$ is the FeatureSE-enhanced state representation. The mode head outputs a Bernoulli distribution over CC and NDA, while the resource head outputs a categorical distribution over candidate budgets using mode-specific representations to avoid cross-mode information leakage. 

\subsubsection{Critic} 

To support the autoregressive policy, we adopt a mode-conditioned critic $V(z,a_{\mathrm{mode}})$ with two value heads $V_0(z)$ and $V_1(z)$. For the resource policy, the chosen-mode value $V(z_t,a_{\mathrm{mode}(t)})$ provides a baseline independent of $a_{\mathrm{res}}$; for the mode policy, we use the mode-marginal baseline
\begin{equation}
    \bar V(z_t)=
    \sum_{m\in\{0,1\}}
    \pi_{\theta_{\mathrm{mode}}}(m\mid z_t)V(z_t,m),
\end{equation}
which is independent of the realized mode action. These baselines reduce variance while preserving unbiased policy-gradient estimation.

\subsection{Workflow}

After training, WCC follows a simple decomposition--decision--execution--update loop. The problem is first decomposed into subspaces using an existing decomposition strategy. At each optimization step, the current state $s_t$ is fed to the agent, which selects the optimization mode (CC or NDA) together with its computational budget. The selected optimizer is then executed under this budget, after which the best solution and state are updated. This process repeats until the evaluation budget is exhausted.

For each policy head $h\in\{\mathrm{mode},\mathrm{res}\}$, the clipped PPO objective is
\begin{equation}
\mathcal{L}_{h}^{\mathrm{clip}}
=
-\mathbb{E}\!\left[
\min\!\left(
\rho_{h,t}A_{h,t},
\operatorname{clip}(\rho_{h,t},1-\epsilon,1+\epsilon)A_{h,t}
\right)
\right],
\end{equation}
where $\rho_{h,t}$ denotes the policy probability ratio and $A_{h,t}$ is estimated by GAE using the corresponding critic baseline. Standard value and entropy regularization are further incorporated during training. Methodological details are provided in Appendix~A.

\section{Experiments}
\label{Experiments}

\subsection{Experimental Setup}

\subsubsection{Comparison algorithms} We instantiate WCC with CMA-ES~\cite{hansen2001completely} as the CC optimizer and MMES~\cite{he2020mmes} as the NDA, denoted as WCC-MMES. We compare it with representative CC-based methods, including DG2~\cite{omidvar2017dg2}, CC-CMAES~\cite{liu2013scaling}, ERDG~\cite{yang2020efficient}, OEDG~\cite{tian2024enhanced}, LCC-CMAES~\cite{guo2025advancing}; NDAs including MMES~\cite{he2020mmes}, sep-CMA-ES~\cite{ros2008simple}, LM-MA-ES~\cite{loshchilov2018large}, FCMAES~\cite{li2018fast}, and MetaES~\cite{lange2023discovering}; and hybrid methods CC-SHADE-ML~\cite{vakhnin2022novel} and HCC-ES~\cite{qiu2025novel}.

\subsubsection{Benchmarks}
The primary benchmark is CEC2013LSGO~\cite{li2013benchmark}, containing 15 problems spanning fully separable, partially separable, overlapping, and non-separable structures. We treat F12 as non-separable because its chain-like structure induces 999 overlapping variables and subspaces. We further use the 1000-dimensional Auto Overlapping Benchmark (AOB)~\cite{qiu2025novel} to evaluate generalization across overlapping and non-separable structures. 

\subsubsection{Settings}
WCC-MMES is trained for 90 epochs on representative training problems with $\mathrm{MaxFEs}=1\mathrm{E}{+06}$ and is directly reused for testing without retraining. Each algorithm is independently evaluated over 30 runs, with comparison methods configured following their original settings. The training and unseen test problems may share the same separability category but differ in their underlying base functions. Detailed experimental settings are provided in Appendix~B.

\begin{table*}[t]
  \centering
   \caption{Comparing WCC-MMES with comparison algorithms on CEC2013LSGO at 1.5E6 FEs.}
  \setlength{\tabcolsep}{6pt}
  \renewcommand{\arraystretch}{1.0}
  \resizebox{\textwidth}{!}{
    \begin{tabular}{r|c|ccccc|ccccc|cc}
    \hline
    \multirow{2}{*}{problem} & \multicolumn{1}{c|}{WCC-based Algorithms}& \multicolumn{5}{c|}{CC-based Algorithms} & \multicolumn{5}{c|}{NDAs} &\multicolumn{2}{c}{Hybrid Algorithms}\\[0.5ex]  \cline{2-14}
     & WCC-MMES & DG2-CMAES & CC-CMAES & LCC-CMAES & ERDG-CMAES & OEDG-CMAES & MMES & sep-CMA-ES & LM-MA-ES & FCMAES & MetaES & HCC-ES & CC-SHADE-SL \\
    \hline

    \multicolumn{1}{c|}{*1}
    & \cellcolor[rgb]{0.88,0.94,1.00}\textbf{2.16E-20}
    & 7.36E-20(+) & 9.12E+06(+) & 1.13E+05(+) & 9.51E-19(+) & 1.15E-18(+)
    & 7.98E+05(+) & 3.90E-17(+) & 4.40E+06(+) & 5.90E+06(+) & 4.49E+11(+)
    & 6.96E-20(+) & 1.87E+05(+) \\
    & \cellcolor[rgb]{0.88,0.94,1.00}{{$\pm$ 4.90E-20}}
    & {{$\pm$ 8.97E-21}} & {{$\pm$ 2.20E+06}} & {{$\pm$ 2.54E+04}}
    & {{$\pm$ 6.35E-19}} & {{$\pm$ 3.47E-19}} & {{$\pm$ 1.37E+05}}
    & {{$\pm$ 7.95E-19}} & {{$\pm$ 3.27E+05}} & {{$\pm$ 8.20E+05}}
    & {{$\pm$ 2.35E+10}} & {{$\pm$ 8.87E-21}} & {{$\pm$ 3.64E+05}} \\
\hline

    \multicolumn{1}{c|}{*4}
    & 2.13E+05
    & \cellcolor[rgb]{0.88,0.94,1.00}\textbf{1.94E+05(-)}
    & 1.18E+11(+) & 3.48E+10(+) & 1.49E+10(+) & 1.87E+10(+)
    & 4.43E+08(+) & 1.38E+11(+) & 7.33E+11(+) & 3.45E+09(+)
    & 2.37E+14(+) & 7.18E+06(+) & 1.25E+10(+) \\
    & {{$\pm$ 4.72E+04}}
    & \cellcolor[rgb]{0.88,0.94,1.00}{{$\pm$ 2.45E+04}}
    & {{$\pm$ 5.01E+11}} & {{$\pm$ 4.99E+09}} & {{$\pm$ 7.32E+09}}
    & {{$\pm$ 5.44E+09}} & {{$\pm$ 7.68E+07}} & {{$\pm$ 2.55E+10}}
    & {{$\pm$ 1.14E+11}} & {{$\pm$ 4.20E+08}} & {{$\pm$ 5.40E+11}}
    & {{$\pm$ 2.85E+06}} & {{$\pm$ 1.10E+10}} \\
\hline

    \multicolumn{1}{c|}{*5}
    & \cellcolor[rgb]{0.88,0.94,1.00}\textbf{1.65E+06}
    & 1.70E+06(+) & 1.05E+07(+) & 1.05E+07(+) & 1.97E+06(+) & 2.33E+06(+)
    & 4.17E+06(+) & 1.67E+06(+) & 1.75E+06(+) & 4.43E+06(+)
    & 1.02E+08(+) & 1.81E+06(+) & 1.92E+06(+) \\
    & \cellcolor[rgb]{0.88,0.94,1.00}{{$\pm$ 3.26E+05}}
    & {{$\pm$ 3.39E+05}} & {{$\pm$ 2.18E+06}} & {{$\pm$ 6.31E+05}}
    & {{$\pm$ 3.09E+05}} & {{$\pm$ 4.04E+05}} & {{$\pm$ 9.25E+05}}
    & {{$\pm$ 1.52E+05}} & {{$\pm$ 1.15E+05}} & {{$\pm$ 8.33E+05}}
    & {{$\pm$ 5.21E+06}} & {{$\pm$ 3.12E+05}} & {{$\pm$ 7.59E+06}} \\
\hline

    \multicolumn{1}{c|}{*8}
    & \cellcolor[rgb]{0.88,0.94,1.00}\textbf{3.97E+09}
    & 4.63E+09(+) & 5.04E+15(+) & 7.50E+14(+) & 4.58E+10(+) & 3.17E+15(+)
    & 4.21E+11(+) & 1.83E+15(+) & 2.90E+15(+) & 3.61E+12(+)
    & 2.49E+18(+) & 6.60E+09(+) & 2.13E+14(+) \\
    & \cellcolor[rgb]{0.88,0.94,1.00}{{$\pm$ 5.21E+8}}
    & {{$\pm$ 1.40E+09}} & {{$\pm$ 7.47E+15}} & {{$\pm$ 2.44E+14}}
    & {{$\pm$ 1.99E+10}} & {{$\pm$ 3.89E+15}} & {{$\pm$ 4.35E+10}}
    & {{$\pm$ 8.79E+14}} & {{$\pm$ 1.47E+15}} & {{$\pm$ 6.78E+11}}
    & {{$\pm$ 1.38E+16}} & {{$\pm$ 1.67E+08}} & {{$\pm$ 2.55E+14}} \\
\hline

    \multicolumn{1}{c|}{*9}
    & 1.70E+08 & 1.84E+08(+) & 7.62E+08(+) & 8.95E+08(+)
    & 1.69E+08($\approx$) & 1.57E+08(-) & 4.41E+08(+)
    & \cellcolor[rgb]{0.88,0.94,1.00}\textbf{1.41E+08(-)}
    & 2.30E+08(+) & 5.70E+08(+) & 1.81E+10(+) & 1.51E+08(-) & 2.34E+08(+) \\
    & {{$\pm$ 5.78E+07}} & {{$\pm$ 2.75E+07}} & {{$\pm$ 2.18E+08}}
    & {{$\pm$ 8.37E+07}} & {{$\pm$ 2.94E+07}} & {{$\pm$ 4.02E+07}}
    & {{$\pm$ 4.26E+07}}
    & \cellcolor[rgb]{0.88,0.94,1.00}{{$\pm$ 1.70E+07}}
    & {{$\pm$ 2.20E+07}} & {{$\pm$ 9.23E+07}} & {{$\pm$ 6.83E+07}}
    & {{$\pm$ 2.76E+07}} & {{$\pm$ 1.13E+08}} \\
\hline

    \multicolumn{1}{c|}{*12}
    & 9.33E+02 & 1.78E+03(+) & 2.27E+03(+) & 1.47E+03(+)
    & 9.58E+02(+) & 2.06E+03(+) & 9.33E+02($\approx$) & 1.00E+03(+)
    & 8.61E+02(-)
    & \cellcolor[rgb]{0.88,0.94,1.00}\textbf{8.44E+02(-)}
    & 9.19E+12(+) & 1.72E+03(+) & 2.10E+05(+) \\
    & {{$\pm$ 8.60E+01}} & {{$\pm$ 4.15E+01}} & {{$\pm$ 7.89E+02}}
    & {{$\pm$ 1.92E+02}} & {{$\pm$ 2.94E+00}} & {{$\pm$ 2.29E+03}}
    & {{$\pm$ 6.74E+01}} & {{$\pm$ 4.22E+01}} & {{$\pm$ 3.41E+01}}
    & \cellcolor[rgb]{0.88,0.94,1.00}{{$\pm$ 1.20E+02}}
    & {{$\pm$ 3.21E+12}} & {{$\pm$ 2.76E+01}} & {{$\pm$ 6.60E+05}} \\
\hline

    \multicolumn{1}{c|}{*13}
    & \cellcolor[rgb]{0.88,0.94,1.00}\textbf{5.56E+06}
    & 9.05E+10(+) & 6.70E+10(+) & 4.33E+08(+) & 1.70E+09(+) & 1.56E+10(+)
    & 9.36E+06(+) & 2.20E+11(+) & 6.76E+07(+) & 2.71E+07(+)
    & 5.74E+18(+) & 2.88E+07(+) & 3.15E+09(+) \\
    & \cellcolor[rgb]{0.88,0.94,1.00}{{$\pm$ 8.57E+05}}
    & {{$\pm$ 3.45E+09}} & {{$\pm$ 3.17E+10}} & {{$\pm$ 2.51E+08}}
    & {{$\pm$ 8.68E+08}} & {{$\pm$ 5.55E+09}} & {{$\pm$ 2.68E+06}}
    & {{$\pm$ 1.13E+11}} & {{$\pm$ 2.15E+06}} & {{$\pm$ 8.49E+06}}
    & {{$\pm$ 7.59E+17}} & {{$\pm$ 2.49E+07}} & {{$\pm$ 1.43E+09}} \\
\hline \hline

    \multicolumn{1}{c|}{2}
    & \cellcolor[rgb]{0.88,0.94,1.00}\textbf{2.06E+03}
    & 3.68E+03(+) & 2.16E+03(+) & 2.11E+03(+) & 1.33E+04(+) & 1.35E+04(+)
    & 1.44E+04(+) & 5.51E+03(+) & 5.82E+03(+) & 1.62E+04(+)
    & 1.15E+05(+) & 3.72E+03(+) & 2.54E+03(+) \\
    & \cellcolor[rgb]{0.88,0.94,1.00}{{$\pm$ 4.53E+02}}
    & {{$\pm$ 4.15E+02}} & {{$\pm$ 1.39E+02}} & {{$\pm$ 1.30E+02}}
    & {{$\pm$ 9.59E+02}} & {{$\pm$ 7.05E+02}} & {{$\pm$ 9.02E+02}}
    & {{$\pm$ 1.59E+02}} & {{$\pm$ 2.74E+02}} & {{$\pm$ 8.52E+02}}
    & {{$\pm$ 5.72E+03}} & {{$\pm$ 3.71E+02}} & {{$\pm$ 1.45E+03}} \\
\hline

    \multicolumn{1}{c|}{3}
    & \cellcolor[rgb]{0.88,0.94,1.00}\textbf{2.00E+01}
    & 2.10E+01(+) & 2.01E+01($\approx$) & 2.04E+01(+) & 2.16E+01(+) & 2.16E+01(+)
    & 2.01E+01($\approx$) & 2.16E+01(+) & 2.16E+01(+) & 2.09E+01(+)
    & 2.16E+01(+) & 2.10E+01(+) & 2.09E+01(+) \\
    & \cellcolor[rgb]{0.88,0.94,1.00}{{$\pm$ 2.31E-02}}
    & {{$\pm$ 3.56E-02}} & {{$\pm$ 2.50E-02}} & {{$\pm$ 4.78E-02}}
    & {{$\pm$ 5.01E-03}} & {{$\pm$ 5.01E-03}} & {{$\pm$ 1.87E-01}}
    & {{$\pm$ 6.06E-03}} & {{$\pm$ 2.67E-03}} & {{$\pm$ 1.10E-01}}
    & {{$\pm$ 7.97E-01}} & {{$\pm$ 3.79E-02}} & {{$\pm$ 8.31E-02}} \\
\hline

    \multicolumn{1}{c|}{6}
    & \cellcolor[rgb]{0.88,0.94,1.00}\textbf{9.95E+05}
    & 1.06E+06(+) & 1.03E+06(+) & 1.05E+06(+) & 1.06E+06(+) & 1.06E+06(+)
    & \textbf{9.95E+05($\approx$)}
    & 1.06E+06(+) & 1.06E+06(+) & 9.96E+05($\approx$)
    & 1.06E+06(+) & 1.06E+06(+) & 1.06E+06(+) \\
    & \cellcolor[rgb]{0.88,0.94,1.00}{{$\pm$ 1.10E+01}}
    & {{$\pm$ 9.01E+02}} & {{$\pm$ 1.58E+04}} & {{$\pm$ 2.00E+03}}
    & {{$\pm$ 1.21E+03}} & {{$\pm$ 1.31E+03}} & {{$\pm$ 4.52E+01}}
    & {{$\pm$ 1.64E+03}} & {{$\pm$ 1.25E+03}} & {{$\pm$ 2.18E+03}}
    & {{$\pm$ 8.23E+03}} & {{$\pm$ 8.28E+02}} & {{$\pm$ 2.42E+03}} \\
\hline

    \multicolumn{1}{c|}{7}
    & 4.41E+04
    & \cellcolor[rgb]{0.88,0.94,1.00}\textbf{3.97E+04(-)}
    & 1.46E+09(+) & 5.36E+07(+) & 5.62E+09(+) & 7.86E+09(+)
    & 7.56E+05(+) & 6.79E+09(+) & 8.23E+06(+) & 4.23E+06(+)
    & 1.09E+18(+) & 1.69E+05(+) & 3.91E+07(+) \\
    & {{$\pm$ 2.37E+04}}
    & \cellcolor[rgb]{0.88,0.94,1.00}{{$\pm$ 1.45E+04}}
    & {{$\pm$ 3.93E+09}} & {{$\pm$ 2.38E+07}} & {{$\pm$ 2.08E+09}}
    & {{$\pm$ 5.61E+09}} & {{$\pm$ 6.72E+04}} & {{$\pm$ 4.62E+09}}
    & {{$\pm$ 9.59E+05}} & {{$\pm$ 5.43E+05}} & {{$\pm$ 7.24E+13}}
    & {{$\pm$ 1.16E+04}} & {{$\pm$ 4.14E+07}} \\
\hline

    \multicolumn{1}{c|}{10}
    & \cellcolor[rgb]{0.88,0.94,1.00}\textbf{9.05E+07}
    & 9.12E+07(+) & 9.42E+07(+) & 9.42E+07(+) & 9.46E+07(+) & 9.50E+07(+)
    & \textbf{9.05E+07($\approx$)}
    & 9.42E+07(+) & 9.42E+07(+) & 9.07E+07(+)
    & 9.44E+07(+) & 9.11E+07(+) & 9.32E+07(+) \\
    & \cellcolor[rgb]{0.88,0.94,1.00}{{$\pm$ 5.09E+02}}
    & {{$\pm$ 2.92E+05}} & {{$\pm$ 4.62E+05}} & {{$\pm$ 4.46E+05}}
    & {{$\pm$ 2.21E+05}} & {{$\pm$ 2.12E+05}} & {{$\pm$ 2.04E+04}}
    & {{$\pm$ 2.08E+05}} & {{$\pm$ 2.91E+05}} & {{$\pm$ 2.83E+05}}
    & {{$\pm$ 3.11E+05}} & {{$\pm$ 4.91E+05}} & {{$\pm$ 3.15E+05}} \\
\hline

    \multicolumn{1}{c|}{11}
    & \cellcolor[rgb]{0.88,0.94,1.00}\textbf{1.45E+07}
    & 5.74E+07(+) & 5.06E+11(+) & 4.81E+10(+) & 1.81E+09(+) & 1.63E+10(+)
    & 1.82E+07(+) & 1.09E+11(+) & 7.61E+08(+) & 1.09E+08(+)
    & 4.88E+19(+) & 5.10E+07(+) & 4.30E+19(+) \\
    & \cellcolor[rgb]{0.88,0.94,1.00}{{$\pm$ 1.63E+06}}
    & {{$\pm$ 4.03E+07}} & {{$\pm$ 4.85E+11}} & {{$\pm$ 3.43E+10}}
    & {{$\pm$ 9.20E+08}} & {{$\pm$ 1.52E+10}} & {{$\pm$ 2.44E+06}}
    & {{$\pm$ 2.90E+10}} & {{$\pm$ 1.39E+08}} & {{$\pm$ 1.91E+07}}
    & {{$\pm$ 8.79E+18}} & {{$\pm$ 2.79E+07}} & {{$\pm$ 6.43E+09}} \\
\hline

    \multicolumn{1}{c|}{14}
    & \cellcolor[rgb]{0.88,0.94,1.00}\textbf{2.52E+07}
    & 1.46E+10(+) & 1.12E+12(+) & 7.31E+08(+) & 2.02E+09(+) & 3.10E+10(+)
    & 3.31E+07(+) & 1.22E+11(+) & 3.27E+08(+) & 1.55E+08(+)
    & 3.41E+23(+) & 8.78E+07(+) & 1.44E+10(+) \\
    & \cellcolor[rgb]{0.88,0.94,1.00}{{$\pm$ 3.74E+06}}
    & {{$\pm$ 2.02E+09}} & {{$\pm$ 3.36E+11}} & {{$\pm$ 5.74E+08}}
    & {{$\pm$ 4.69E+08}} & {{$\pm$ 5.07E+10}} & {{$\pm$ 5.57E+06}}
    & {{$\pm$ 2.17E+11}} & {{$\pm$ 3.44E+07}} & {{$\pm$ 7.54E+06}}
    & {{$\pm$ 7.81E+21}} & {{$\pm$ 1.15E+07}} & {{$\pm$ 2.74E+10}} \\
\hline

    \multicolumn{1}{c|}{15}
    & \cellcolor[rgb]{0.88,0.94,1.00}\textbf{2.31E+06}
    & 6.32E+13(+) & 3.45E+08(+) & 3.78E+07(+) & 4.55E+08(+) & 4.56E+08(+)
    & 5.89E+06(+) & 6.63E+08(+) & 2.31E+07(+) & 1.75E+07(+)
    & 1.26E+18(+) & 4.48E+13(+) & 1.22E+07(+) \\
    & \cellcolor[rgb]{0.88,0.94,1.00}{{$\pm$ 1.47E+06}}
    & {{$\pm$ 4.48E+13}} & {{$\pm$ 3.08E+08}} & {{$\pm$ 4.04E+06}}
    & {{$\pm$ 3.74E+07}} & {{$\pm$ 3.82E+07}} & {{$\pm$ 9.36E+05}}
    & {{$\pm$ 1.26E+08}} & {{$\pm$ 6.26E+06}} & {{$\pm$ 2.31E+06}}
    & {{$\pm$ 6.94E+14}} & {{$\pm$ 2.41E+13}} & {{$\pm$ 5.76E+06}} \\
\hline

    \multicolumn{1}{c|}{+/$\approx$/-}
    & NA & 13/0/2 & 14/1/0 & 15/0/0 & 14/1/0 & 14/0/1
    & 11/4/0 & 14/0/1 & 14/0/1 & 13/1/1 & 15/0/0 & 14/0/1 & 15/0/0 \\
    \hline
    \end{tabular}%
    }

\noindent \parbox{\textwidth}{\footnotesize
Problems marked with * are representative problems used to form the training set.\\
The symbols ‘+’, ‘$-$’, and ‘$\approx$’ signify that WCC-MMES is significantly better than, worse than, or comparable to its competitor, respectively.}

  \label{comparison_algorithms1}
\end{table*}%

\subsection{Comparison Analysis}
\label{Comparison Analysis}
\subsubsection{Optimization Performance Comparison}
\label{Optimization Performance Comparison}
Table~\ref{comparison_algorithms1} reports results on CEC2013LSGO at $\mathrm{MaxFEs}=1.5\mathrm{E}{+06}$ (including $5\mathrm{E}{+05}$ for DG2). Symbols “+”, “$-$”, and “$\approx$” indicate significantly better, worse, and comparable performance, respectively (Wilcoxon rank-sum test, $\alpha=0.05$), with W/T/L counts summarized in the final row.

Based on the results from Table~\ref{comparison_algorithms1}, we can analyze the following outcomes: 
\textbf{a) WCC demonstrates superior performance:} WCC-MMES consistently performs competitively against all other algorithms. In particular, DG2-CMAES achieves theoretically maximal decomposition accuracy, and MMES is among the most competitive NDAs. WCC’s strong performance thus highlights the power of hybridizing the two approaches.
\textbf{b) Data-driven approaches transcend the limitations of expert heuristics:}
    With expert knowledge scarce, WCC-MMES makes finer-grained action selections in a data-driven manner. Its superior performance over hybrid algorithms further shows that WCC can transcend the limitations of expert heuristics.
\textbf{c) Effectively Learning the Strengths of CC and NDAs:} WCC-MMES consistently excels on problems favoring both CC-based algorithms (e.g., F1, F5) and MMES (e.g., F13–F15), demonstrating its ability to effectively integrate the strengths of CC and NDAs.

\subsubsection{Generalization Experiment}
\label{Generalization Experiment}

We evaluate WCC-MMES generalization across three settings by directly testing the trained policy without retraining.

\begin{figure}[t]
    \centering
    \includegraphics[width=0.8\linewidth]{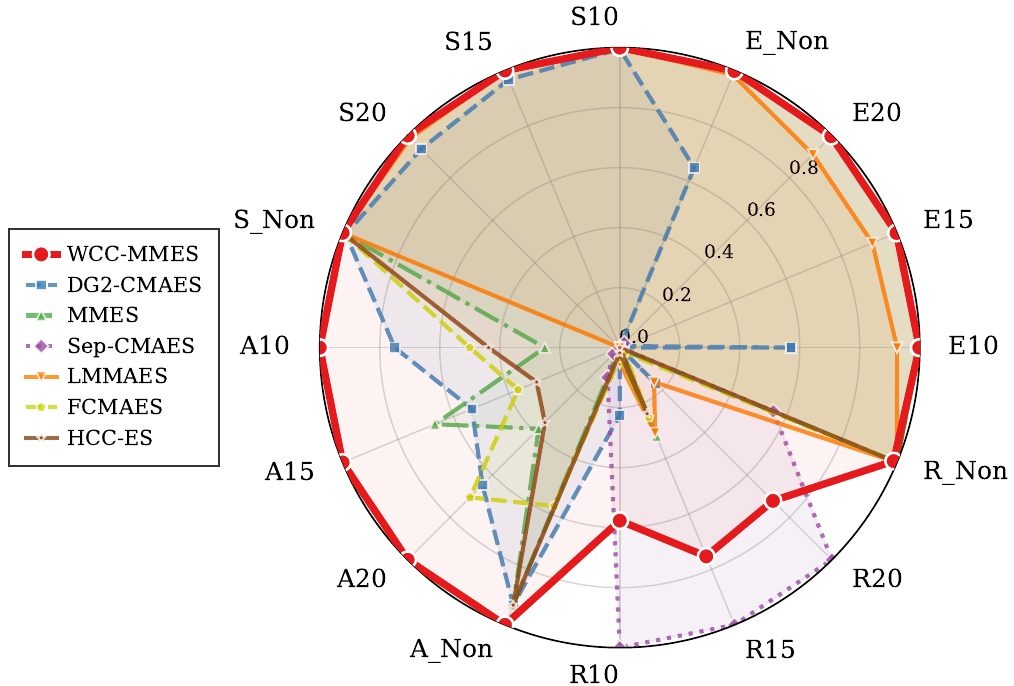}
    \caption{\textcolor{black}{Comparison of WCC-MMES and baselines on AOB at $1.5$E$6$ FEs. 
Radial values are min--max normalized scores, with larger values indicating better relative performance.}}
    \label{AOB 1E6FEs}
\end{figure}

\paragraph{Generalization across Problems}
In Table \ref{comparison_algorithms1}, 8 of the 15 test problems were unseen during training, yet WCC-MMES remained competitive; notably, its strong performance on F15—a complex non-separable problem—underscores its promising generalization ability. To further validate WCC-MMES’s competitiveness on overlapping and non-separable problems and its generalization across problem instances, we generated 16 new AOB problems and tested them under MaxFEs = $1.5 \text{E+06}$. Figure \ref{AOB 1E6FEs} shows that WCC-MMES is competitive on most problems, outperforming both MMES and DG2-CMAES. Its strength on non-separable problems is particularly notable; for example, in the Rastrigin family, WCC-MMES lags on overlapping variants but still excels on non-separable instances, further confirming its ability to tackle complex non-separable problems. 

\paragraph{Generalization across Optimization Horizons}

Although trained with only $\mathrm{MaxFEs}=1\mathrm{E}{+06}$, WCC-MMES remains highly competitive when directly evaluated at $3\mathrm{E}{+06}$ FEs without retraining. It achieves +/-/$\approx$ records of 13/0/2, 15/0/0, 12/3/0, and 14/0/1 against DG2-CMAES, LCC-CMAES, MMES, and HCC-ES, respectively. The advantage is also substantial on several problems; for example, on F15, WCC-MMES reduces the final cost from the order of $10^{13}$ achieved by DG2-CMAES and HCC-ES to $10^{5}$, corresponding to nearly eight orders of magnitude improvement. These results demonstrate strong generalization across optimization horizons. Complete results are provided in Appendix~B.

\begin{figure}[t]
    \centering
    \includegraphics[width=\linewidth]{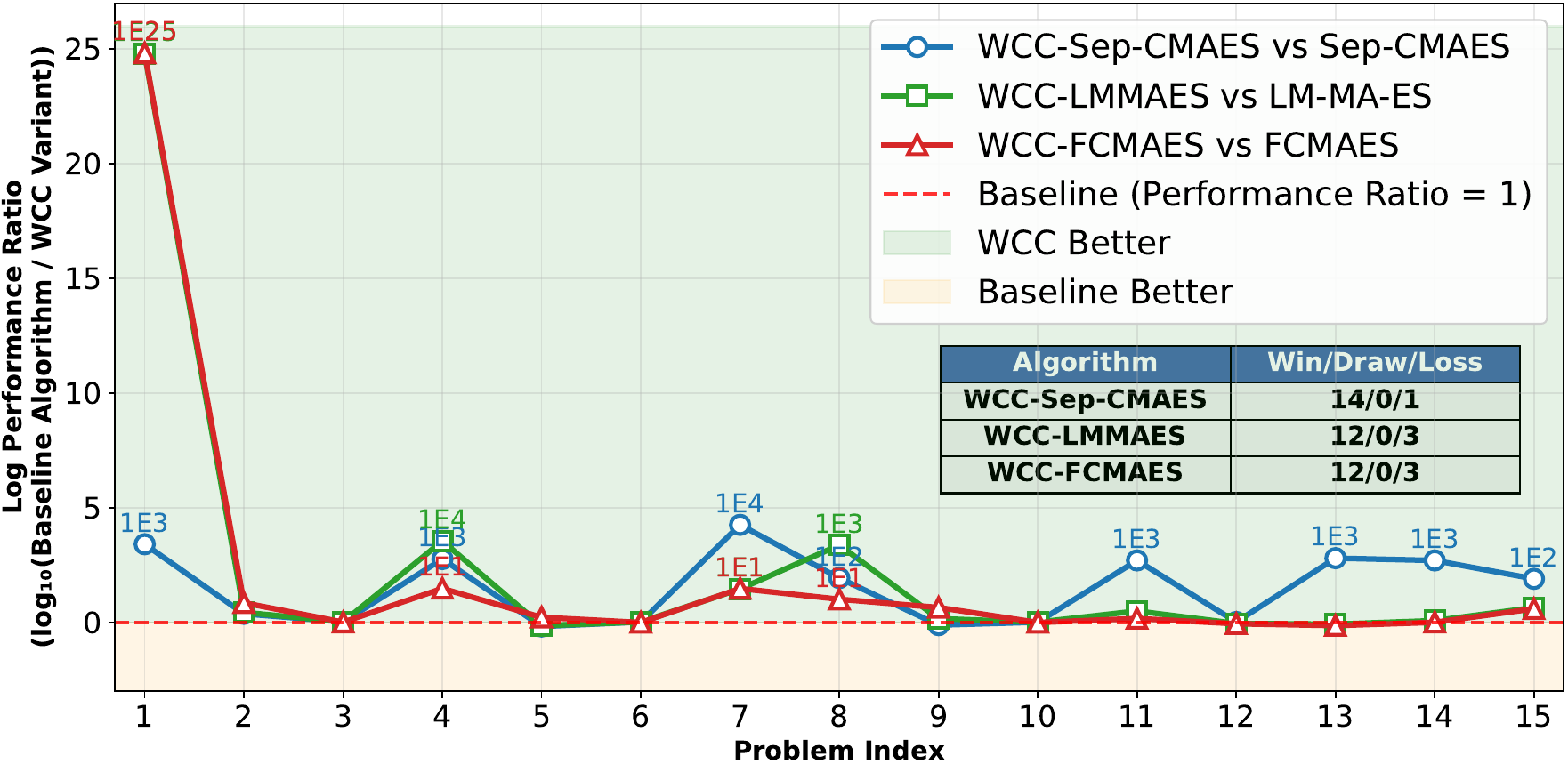}
    \caption{Generalization experiment of WCC across baseline algorithms.}
    \label{Generality}
\end{figure}

\paragraph{Generalization across Baseline Algorithms}
\label{Generality Experiments}
To assess whether WCC generalizes beyond MMES, we directly replace the NDA component in the trained WCC-MMES with Sep-CMAES, LM-MA-ES, and FCMAES, yielding three WCC variants without retraining. As shown in Figure~\ref{Generality}, these variants outperform their corresponding standalone NDAs on most CEC2013LSGO problems at $\mathrm{MaxFEs}=1\mathrm{E}{+06}$, often by large margins. This transferability stems from WCC's optimizer-agnostic state and action design, which preserves the same meta-level mode-selection and budget-allocation problem across different NDAs.

\subsection{Ablation Study}
\label{Ablation Study}

\textcolor{black}{To further examine where the performance of WCC-MMES comes from, we conduct a comprehensive ablation study from four aspects: action space, state space, reward function, and network architecture. }

\subsubsection{Experimental Protocol}
We ablate WCC from four aspects: action, state, reward, and network design. For the action space, we randomize or fix either the mode or resource decision to examine the necessity of jointly learning both components. For the state space, each of the four feature groups \{$\mathcal{S}_{pro}$, $\mathcal{S}_{conv}$, $\mathcal{S}_{res}$,  $\mathcal{S}_{mode}$\} is removed individually. For the reward, the proposed log-ratio formulation is replaced with Difference, Sparse Sign, and Linear alternatives. For the network, we remove FeatureSE, replace the autoregressive policy with a simple MLP, or remove both. All variants follow the same training and evaluation protocol, and their final costs are min--max normalized within each category before averaging across problems. 

\subsubsection{Ablation Analysis}
\begin{figure}[t]
    \centering
    \includegraphics[width=\linewidth]{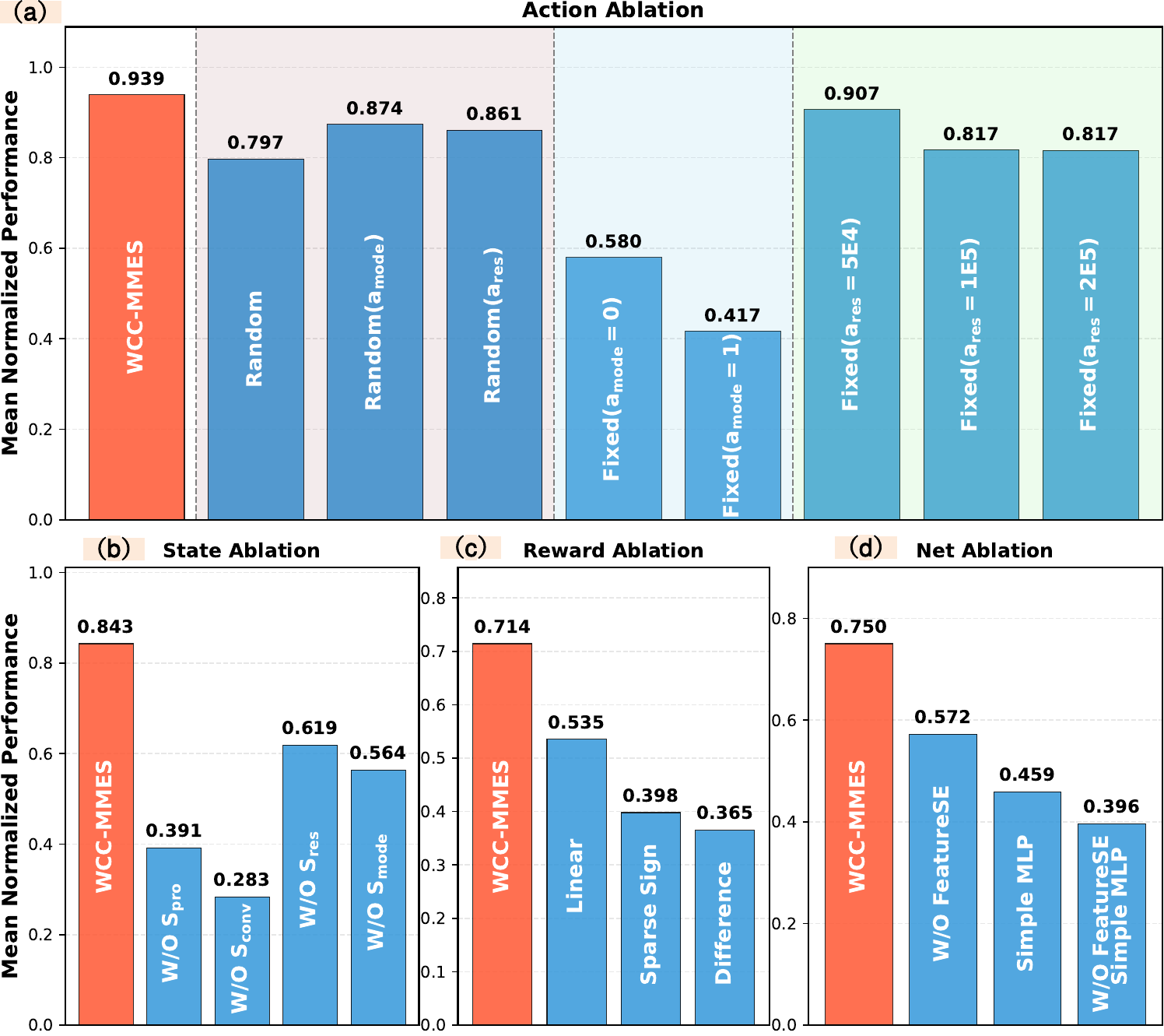}
    \caption{\textcolor{black}{Ablation study, including (a) action space ablation, (b) state space ablation, (c) reward function ablation, and (d) network architecture ablation.}}
    \label{Ablation Study Fig}
\end{figure}

Figure~\ref{Ablation Study Fig} confirms the contribution of all four design components. For the action space, randomizing or fixing either decision degrades performance, showing the importance of jointly learning mode selection and resource allocation. For the state space, removing any feature group causes a clear drop, with convergence features having the largest impact, indicating that the four groups provide complementary information. For the reward, all alternative formulations underperform the proposed log-ratio reward, highlighting the importance of scale-robust and numerically stable feedback. For the network, removing FeatureSE or replacing the autoregressive policy with a simple MLP consistently worsens performance, with the largest degradation occurring when both are removed, validating the joint roles of adaptive state reweighting and mode-conditioned resource allocation. Detailed ablation configurations and further analysis are provided in Appendix~B.

\begin{figure*}[t]
    \centering
    \includegraphics[width=\linewidth]{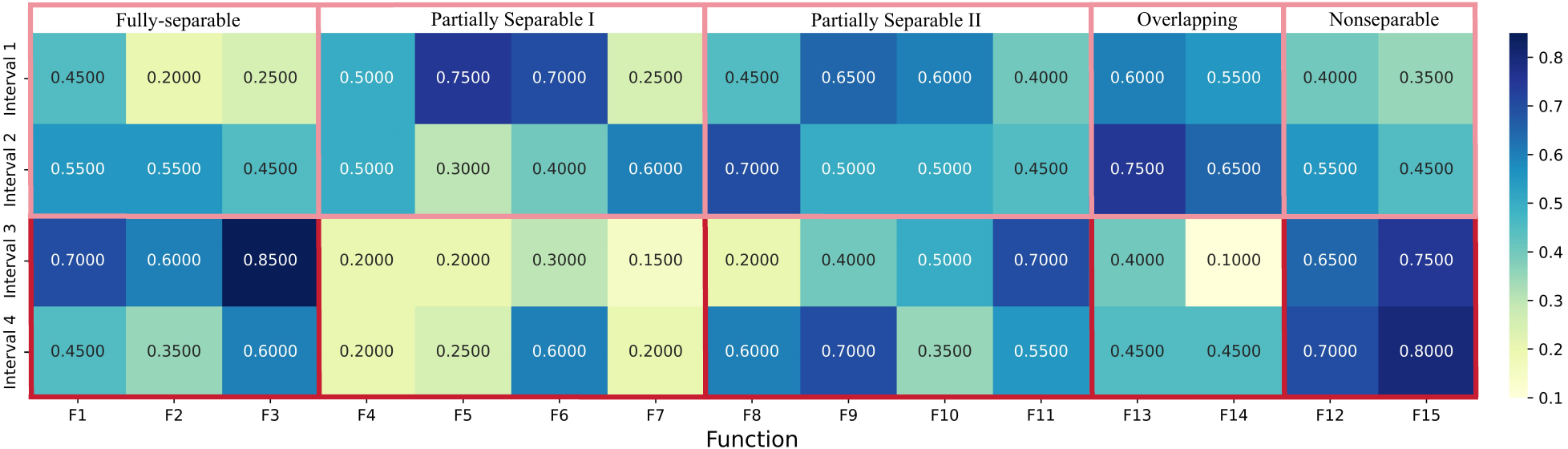}
    \caption{\textcolor{black}{Proportion of NDA selections across problem types and optimization stages. 
Each cell reports the mean proportion of $a_{\mathrm{mode}}=\mathrm{NDA}$, where darker colors indicate more frequent NDA usage. 
Rows denote four equal intervals divided by MaxFEs, and columns denote test functions grouped by separability.}}
    \label{select_cc}
\end{figure*}

\subsection{Further In-depth Analysis}

\label{Analysis of When WCC Chooses CC}

Understanding when WCC selects CC is a central focus of this work. We divide each optimization run into four equal intervals and report the mean proportion of NDA selections in Figure~\ref{select_cc}, revealing clear patterns across both problem structures and optimization stages.

Across problem structures, WCC increasingly favors NDAs as variable coupling becomes stronger. CC dominates on fully and partially separable problems by exploiting decomposable structures, whereas NDA usage increases on overlapping problems and becomes dominant on non-separable ones, where decomposition is less effective. Across optimization stages, WCC generally favors NDAs early and gradually shifts toward CC on separable and overlapping problems; in contrast, on non-separable problems, CC is used mainly for limited early exploration, while NDAs dominate the middle and later stages. These results show that the appropriate use of CC depends jointly on problem structure and optimization stage rather than on a fixed switching rule.

Individual functions may nevertheless deviate from these category-level trends. For example, although F3 is categorized as fully separable, its exact separability is preserved only under one-dimensional decomposition; the 50-dimensional grouping used here therefore limits the effectiveness of CC, causing WCC to increasingly favor NDAs in later stages, similarly to non-separable problems. A similar pattern appears on F6. This further indicates that WCC adapts its mode selection to the observed optimization context rather than relying solely on nominal separability.

\section{CONCLUSION AND FUTURE WORK}
\label{CONCLUSION AND FUTURE WORK}
We proposed WCC, a pioneering framework that learns when to use CC in LSGO, concurrently optimizing the resource allocation. WCC enabled fine-grained, dynamic selection and effective hybridization of CC and NDAs without human expertise. Extensively evaluated, WCC exhibited strong competitiveness—especially on non-separable problems, where both CC and NDAs alone typically struggle. Moreover, WCC demonstrated generalization across problems, optimization horizons, and baseline optimizers, enabling train-once under a low FE budget and reuse with different underlying optimizers to mitigate LSGO’s computational cost. Finally, ablation studies and in-depth analyses yielded insights that go beyond existing theoretical understanding, offering practical guidance on when and how to deploy CC. 

By shifting the focus from how to decompose to when to deploy CC, WCC opens a new direction for CC research. Future work will explore online separability characterization and adaptive selection among multiple CC and NDA strategies.

\bibliographystyle{IEEEtran}
\bibliography{ref}

\end{document}


\title{Appendix}
\author{}
\maketitle
\appendices
\providecommand{\wccappendixroot}{.}
\providecommand{\mainfig}[1]{Fig.~#1 of the main paper}
\providecommand{\mainequation}[1]{Equation~(#1) of the main paper}
\section{Methodology}
\label{supp:methodology}
\subsection{State Representation}
\label{supp:state}
WCC represents the optimization context with 16 features, grouped into
problem ($s_1$--$s_5$), convergence ($s_6$--$s_{11}$), resource
($s_{12}$--$s_{14}$), and mode-history ($s_{15}$--$s_{16}$) information.
Table~\ref{supp:State-features} lists their complete definitions.
Here $D$ is the problem dimension, $\Theta$ is the dependency structure
matrix (DSM), $\mathbb{S}$ is the subspace list, and $\Omega_i$ is the
variable set of subspace $i$. The best-so-far objective value is denoted by
$cost_t^*$, while $sumFEs$ counts consumed function evaluations.
The following definitions expand the trajectory-based and mode-history
features that require more than a single expression.
\begin{table*}[tp]
\centering
\caption{Complete definitions of the 16 WCC state features.}
\label{supp:State-features}
\footnotesize
\begin{tabularx}{\textwidth}{@{}c c p{0.35\textwidth} X@{}}
\hline
Group & Index & Formula & Explanation \\
\hline
$\mathcal{S}_{pro}$ & 1 & $ \frac{1000}{D}$ & The relative size of the problem dimensions $D$. \\
 & 2 & $ 1-\frac{\sum_{i=1}^{D} \sum_{j=1}^{D} \mathbb{I}\left(\Theta[i,j]=0 \right)}{D^2}$ & The sparsity of DSM $\Theta$ characterizes the interactive complexity of the problem. \\
 & 3 & $ \frac{\left|\{i| \sum_{j=1}^{D}\Theta[i,j]=1\}\right|}{D}$ & The proportion of completely separable variables. \\
 & 4 & $ \frac{\left|\bigcup_{i=1}^{|\mathbb{S}|} \bigcup_{j \neq i}^{|\mathbb{S}|}\left(\Omega_{i} \cap \Omega_{j}\right)\right|}{D}$ & The proportion of overlapping variables, $\Omega_{i}$ is the variable set of subspace $i$. \\
 & 5 & $ \frac{|\mathbb{S}|}{D}$ & The relative size of the number of subspaces $|\mathbb{S}|$. \\
\hline
$\mathcal{S}_{conv}$ & 6 & $ \frac{cost^*_{0}-cost^*_{t-1}}{cost^*_{0}}$ & The decrease ratio of optimal cost $cost_{t-1}^*$ at step $t-1$. \\
 & 7 & $LFER(cost^*)$ & The proportion of low-frequency energy (LFER) in the optimal cost curve. \\
 & 8 & $-\tanh\!\left(\lambda_1 \cdot \mathrm{mean}(\nabla cost^*)\right)$ & The normalized average first-order trend of the optimal cost trajectory, reflecting the convergence speed. \\
 & 9 & $-\tanh\!\left(\lambda_2 \cdot \mathrm{mean}(\nabla^2 cost^*)\right)$ & The normalized second-order trend indicating convergence acceleration. \\
 & 10 & $\tanh\!\left(\lambda_3
                                    \frac{\mathrm{IQR}(\{\Delta cost_k\}_{k=\tau_0}^{t-1})}
                                    {\left|\mathrm{Median}(\{\Delta cost_k\}_{k=\tau_0}^{t-1})\right|+\epsilon}
                                    \right)$ & The robust variation coefficient of recent optimal-cost changes, measuring the stability of recent convergence. \\
 & 11 & $\frac{dmax_{t-1}}{diameter}$ & The ratio of the maximum distance of individuals  $dmax$ to the search diameter $diameter$ at step $t-1$. \\
\hline
$\mathcal{S}_{res}$ & 12 & $\frac{\text{MaxFEs}-sumFEs}{\text{MaxFEs}}$ & The remaining proportion of FEs. \\
 & 13 & $\frac{mean(FEs_{NDA})}{max(a_{res})}$ & The mean of function evaluations consumed by NDA. \\
 & 14 & $\frac{mean(FEs_{CC})}{max(a_{res})}$ & The mean of function evaluations consumed by CC. \\
\hline
$\mathcal{S}_{mode}$ & 15 & $\frac{\sum_{t=1}^{T}\mathbb{I}(a_t=NDA)}{T}$ & The selection ratio of NDA across all optimization steps $T$. \\
 & 16 & $\bar r_{\mathrm{NDA}}-\bar r_{\mathrm{CC}}$ & The difference between the average recent fitness reductions achieved by NDA and CC. \\
\hline
\end{tabularx}
\end{table*}

\subsubsection{Low-frequency energy ratio ($s_7$)}
For an observed best-cost sequence $\{cost_k^*\}_{k=0}^{T-1}$, compute
the discrete Fourier transform and its normalized amplitude:
\begin{equation}
\hat C_k=\sum_{u=0}^{T-1}cost_u^*e^{-\mathrm{i}2\pi ku/T},
\qquad A_k=\frac{|\hat C_k|}{T}.
\end{equation}
Retain the positive-frequency indices
$\mathcal{K}=\{0,\ldots,\lfloor T/2\rfloor-1\}$.
With $f_k=k/T$ and cutoff $f_c=0.1$, define
\begin{equation}
\begin{aligned}
E_{low}&=\sum_{\substack{k\in\mathcal{K}\\ f_k\le f_c}}A_k,
\qquad E_{tot}=\sum_{k\in\mathcal{K}}A_k,\\
s_7&=\begin{cases}E_{low}/E_{tot},& E_{tot}>0,\\0,& E_{tot}=0.\end{cases}
\end{aligned}
\end{equation}
The ratio summarizes the low-frequency spectral mass of the observed cost
trajectory; the historical definition sums amplitudes.

\subsubsection{First-order convergence trend ($s_8$)}
For the available trajectory $\{cost_k^*\}_{k=0}^{t-1}$, let
$g_k=\nabla cost_k^*$ be its numerical gradient and
$\bar g=\operatorname{mean}(\{g_k\}_{k=0}^{t-1})$.
The feature is
\begin{equation}
s_8=-\tanh(\lambda_1\bar g),\qquad \lambda_1=0.5.
\end{equation}
The sign convention makes decreasing costs correspond to positive values,
while the hyperbolic tangent limits the scale of the trend statistic.

\subsubsection{Second-order convergence trend ($s_9$)}
For interior points, use
\begin{equation}
h_k=cost_{k+1}^*-2cost_k^*+cost_{k-1}^*,
\quad k=1,\ldots,t-2.
\end{equation}
The two boundary expressions are
\begin{equation}
\begin{aligned}
h_0&=2cost_0^*-5cost_1^*+4cost_2^*-cost_3^*,\\
h_{t-1}&=2cost_{t-1}^*-5cost_{t-2}^*
          +4cost_{t-3}^*-cost_{t-4}^*.
\end{aligned}
\end{equation}
With $\bar h=\operatorname{mean}(\{h_k\}_{k=0}^{t-1})$, set
\begin{equation}
s_9=-\tanh(\lambda_2\bar h),\qquad \lambda_2=0.5.
\end{equation}
These boundary formulas require at least four trajectory values.

\subsubsection{Robust variation of cost changes ($s_{10}$)}
Use the most recent window of length $L=128$, with
$\tau_0=\max(1,t-L)$, and define
$\Delta cost_k=cost_k^*-cost_{k-1}^*$ for
$k=\tau_0,\ldots,t-1$.
Let $med$ be the median of these changes and
$IQR=Q_{0.75}-Q_{0.25}$ their interquartile range. Then
\begin{equation}
s_{10}=\tanh\!\left(\lambda_3\frac{IQR}{|med|+\epsilon}\right),
\quad \lambda_3=0.5,\quad \epsilon=10^{-12}.
\end{equation}
When fewer than $L$ changes are available, use all available changes.
The feature summarizes fluctuation relative to the typical recent progress.

\subsubsection{Recent mode-specific reductions ($s_{16}$)}
Record the fitness reductions obtained by NDA and CC in separate histories
$\mathcal{R}_{NDA,t}$ and $\mathcal{R}_{CC,t}$ before decision step $t$.
For either mode $m\in\{NDA,CC\}$, define
\begin{equation}
\bar r_m=\begin{cases}
\operatorname{mean}(\mathcal{R}_{m,t}^{[-5:]}),&|\mathcal{R}_{m,t}|>5,\\
\operatorname{mean}(\mathcal{R}_{m,t}),&0<|\mathcal{R}_{m,t}|\le5,\\
0,&|\mathcal{R}_{m,t}|=0.
\end{cases}
\end{equation}
The feature $s_{16}=\bar r_{NDA}-\bar r_{CC}$ compares the average
reductions over the latest five uses of each mode, or over all its uses
when fewer than five are available. Positive values indicate larger recent
reductions from NDA.
\subsection{Network Architecture}
\label{supp:network}
This subsection expands the architecture in \mainfig{1}.
FeatureSE reweights heterogeneous state information; the autoregressive
actor conditions resource allocation on the selected mode; and the critic
provides a corresponding baseline for each policy head.

\subsubsection{Feature-aware state encoding}
Apply layer normalization to obtain
$\tilde{\mathcal{S}}=\operatorname{LN}(\mathcal{S})\in\mathbb{R}^{16}$.
For each of the four feature groups, compute its mean statistic and collect
these statistics in
$\bar{\mathcal{S}}=[\bar{\mathcal{S}}_{pro},\bar{\mathcal{S}}_{conv},
\bar{\mathcal{S}}_{res},\bar{\mathcal{S}}_{mode}]$.
The group and feature gates are
\begin{equation}
\begin{aligned}
g&=\sigma\bigl(\operatorname{MLP}_g(\bar{\mathcal{S}})\bigr),
&g&\in(0,1)^4,\\
h&=\sigma\bigl(\operatorname{MLP}_h(\tilde{\mathcal{S}})\bigr),
&h&\in(0,1)^{16}.
\end{aligned}
\end{equation}
The two MLPs have dimensions $4\!\times\!8\!\times\!4$ and
$16\!\times\!8\!\times\!16$, respectively, with SiLU nonlinearities
and sigmoid outputs. Following \mainequation{4}, residual gating uses
\begin{equation}
w_j=(1-\lambda)+\lambda g_{k(j)}h_j,
\qquad z_j=s_jw_j,\qquad \lambda=0.2.
\end{equation}
Thus $w_j\in(0.8,1)$, limiting feature suppression. The last-layer biases
of both gating MLPs are initialized to 2.0, for which
$\sigma(2)\approx0.88$, to start close to the identity mapping.

\subsubsection{Autoregressive actor}
The policy factorization is
\begin{equation}
\pi(a_{mode},a_{res}\mid z)
=\pi_{\theta_{mode}}(a_{mode}\mid z)
 \pi_{\theta_{res}}(a_{res}\mid z,a_{mode}).
\end{equation}
The mode head is an MLP with dimensions
$16\!\times\!128\!\times\!256\!\times\!1$, ReLU hidden activations,
and a Bernoulli output. Mode 0 denotes NDA and mode 1 denotes CC.
The resource policy uses two separate MLPs, one for each mode, with
dimensions $15\!\times\!160\!\times\!320\!\times\!3$ and ReLU hidden
activations. Each resource head uses a 15-dimensional input excluding the
opposite mode's historical FE statistic: $z^{(0)}$ omits $z_{14}$ for NDA,
while $z^{(1)}$ omits $z_{13}$ for CC. Each head produces a categorical
distribution over the three resource levels in Table~\ref{supp:tab:settings}.
The resource logits for both modes are computed in parallel; the sampled
mode determines which logits are used to sample the resource action.

\subsubsection{Mode-conditioned critic}
The critic uses a backbone of the same style as the actor's mode head,
followed by separate linear value heads $V_0(z)$ and $V_1(z)$.
It does not share actor parameters. The chosen-mode value is
\begin{equation}
\label{supp:eq16}
V(z_t,a_{mode(t)})=\begin{cases}
V_0(z_t),&a_{mode(t)}=0,\\
V_1(z_t),&a_{mode(t)}=1.
\end{cases}
\end{equation}
For the resource head, this value is independent of the sampled resource
action, conditional on the selected mode. For the mode head, the baseline is
\begin{equation}
\label{supp:eq17}
\bar V(z_t)=\sum_{m\in\{0,1\}}\pi_{\theta_{mode}}(m\mid z_t)V(z_t,m),
\end{equation}
which is independent of the realized mode action. This distinction matches
the two successive decisions in the actor and reduces policy-gradient
variance without introducing action-dependent baselines for either head.
\subsection{Training Objectives}
\label{supp:training}
We train the agent $\pi_{\theta}$ by PPO \cite{schulman2017proximal}. At each step $t$, let $d_t\in \{0,1\}$ be the done flag. The auto-regressive policy has two heads $h\in \{\text{mode},\text{res}\}$ with distributions $\pi_{\theta_h}(\cdot)$. Using the mode-marginal value in Equation (\ref{supp:eq17}), the one-step target is
$R_{\text{cond}(t)}=r_{t}+\gamma (1-d_t) \bar{V}(z_{t+1})$.
With value clipping $V_\text{clip}$ around the $V^{\text {old }}$, the value loss is:
\begin{equation}
    \mathcal{L}_V = \mathbb{E}\left[\max\left((V - R_{\text{cond}})^2, (V_{\text{clip}} - R_{\text{cond}})^2\right)\right].
\end{equation}
Advantages $A_{h,t}$ are computed with Generalized Advantage Estimation using the corresponding baseline (standardized per batch). Store old log-probabilities $\log \pi_{\theta_{h}}^{\text {old }}(\cdot)$ and recompute log-probabilities under the current policy $\log \pi_{\theta_{h}}^{\text {new}}(\cdot)$ (no resampling) to form
\begin{equation}
\rho_{h, t}=\exp (\log \pi_{\theta_{h}}^{new}\left(a_{h, t} \mid \cdot\right)-\log \pi_{\theta_{h}}^{\text {old}}\left(a_{h, t} \mid \cdot\right)).
\end{equation}
The clipped surrogate per head is
\begin{equation}
    \mathcal{L}_{h}^{\mathrm{clip}}=-\mathbb{E}\left[\min \left(\rho_{h} A_{h}, \operatorname{clip}\left(\rho_{h}, 1-\varepsilon, 1+\varepsilon\right) A_{h}\right)\right].
\end{equation}
The total loss is
\begin{equation}
\begin{split}
\mathcal{L}={}&\mathcal{L}_{\text{mode }}^{\text {clip }}+\mathcal{L}_{\text{res }}^{\text {clip }}+c_{V} \mathcal{L}_{V}\\
&-\sum_{h \in\{\text {mode, } \mathrm{res}\}} \beta_{h} \mathbb{E}\left[\mathcal{H}\left(\pi_{\theta_{h}}\right)\right]
\end{split}
\end{equation}
where $c_{V}$  weights the value loss and $\beta_{\text {h }}$ controls per-head entropy regularization.
We optimize with Adam (gradient clipping; exponentially decayed learning rate). Each iteration collects $n_{\text{step}}$ transitions per environment, then performs $K$ epochs over the full segment (masked by ($1-d_t$)).
\subsection{Workflow}
\label{supp:workflow}

\begin{algorithm}[tbp]
    \caption{WCC}
    \label{supp:wcc}
    \renewcommand{\algorithmicrequire}{\textbf{Input:}}
    \renewcommand{\algorithmicensure}{\textbf{Output:}}
    \begin{algorithmic}[1]
        \REQUIRE Problem instance $f$ and info $I$, Pre-trained Agent $\pi^{*}_{\theta}$.
        \ENSURE The best individual $X_{best}$, The best cost $Y_{best}$.
        \STATE DSM $\Theta \gets \text{Decomposition Strategy}(f, I)$
        \STATE The list of Subspaces $\mathbb{S} \gets \text{RDDSM} (\Theta)$ 
        \STATE Initialize $X_{best}$ , $Y_{best}$ and state vector $s_0$
        \WHILE {Termination conditions not met} 
            \STATE $a_{mode(t)}$, $a_{res(t)} \gets$  $\pi^{*}_{\theta}(s_{t})$
            \IF{$a_{mode(t)}$ = 0}
                \STATE NDA($f$, $I$, $X_{best}$, $a_{res(t)}$)
                \STATE $X_{best}$, $Y_{best}$, $s_{t+1} \gets $  $\text{Update} \hspace{0.1em}$ $X_{best}$, $Y_{best}$, $s_{t}$
            \ELSE
                \STATE $\text{Each subspace FEs } SubFEs = \left \lceil a_{res(t)}/len(\mathbb{S}) \right \rceil $
                \FOR{$sub$ in $\mathbb{S}$}
                    \STATE $A_{sub}$($f$, $I$, $sub$, $X_{best}$, $SubFEs$)
                    \STATE $X_{best}$, $Y_{best}$ $\gets $  $\text{Update} \hspace{0.1em}$ $X_{best}$, $Y_{best}$
                \ENDFOR
                \STATE $X_{best}$, $Y_{best}$, $s_{t+1} \gets $  $\text{Update} \hspace{0.1em}$ $X_{best}$, $Y_{best}$, $s_{t}$
            \ENDIF
        \ENDWHILE
        \RETURN $X_{best}$, $Y_{best}$
    \end{algorithmic}
\end{algorithm}

Algorithm~\ref{supp:wcc} preserves the complete inference procedure. A decomposition strategy obtains the DSM $\Theta$, and RDDSM derives the subspace list $\mathbb{S}$. The trained agent selects the mode and resource budget from the current state. NDA operates in the full space; CC divides the selected budget among the subspaces and updates the incumbent after each subspace. The state is updated after the selected mode finishes, and optimization stops when no FEs remain.
\section{Experiments}
\label{supp:experiments}
\subsection{Experimental Setup}
\label{supp:setup}
\subsubsection{Benchmarks}
CEC2013LSGO~\cite{li2013benchmark} contains 15 problems, grouped here as
fully separable (F1--F3), partially separable with separable components
(F4--F7), partially separable without separable components (F8--F11),
overlapping (F13--F14), and non-separable (F12 and F15).
As in the main paper, F12 is treated as non-separable because its
chain-like dependency structure yields 999 overlapping variables and
subspaces.

The Auto Overlapping Benchmark (AOB)~\cite{qiu2025novel} provides
additional 1000-dimensional problems while keeping the dimension fixed as
overlap changes. Its base functions are Ackley, Rastrigin, Schwefel, and
Elliptic, following the F8--F11 settings of CEC2013LSGO. The 16 instances
shown in \mainfig{2} comprise three overlapping variants per base function,
with 10, 15, or 20 overlapping variables per subspace, and one globally
rotated non-separable variant. Labels such as A10 indicate the base
function and overlap; A\_Non denotes the non-separable Ackley variant.

\subsubsection{Comparison algorithms}
WCC-MMES uses CMA-ES~\cite{hansen2001completely} for subspace
optimization and MMES~\cite{he2020mmes} for full-space optimization.
DG2 and RDDSM provide the decomposition and subspace construction
used by its CC component~\cite{omidvar2017dg2,qiu2025novel}.
The CC-based comparisons are DG2-CMAES, CC-CMAES~\cite{liu2013scaling},
LCC-CMAES~\cite{guo2025advancing}, ERDG-CMAES~\cite{yang2020efficient},
and OEDG-CMAES~\cite{tian2024enhanced}.
The NDA comparisons are MMES, sep-CMA-ES~\cite{ros2008simple},
LM-MA-ES~\cite{loshchilov2018large}, FCMAES~\cite{li2018fast}, and
MetaES~\cite{lange2023discovering}.
HCC-ES~\cite{qiu2025novel} and CC-SHADE-ML~\cite{vakhnin2022novel}
are the hybrid comparisons. The original numerical table labels the latter
CC-SHADE-SL; that table label is retained below to match Table~I of the
main paper.

\subsubsection{Training and evaluation}
Table~\ref{supp:tab:settings} collects the reported settings.
The training functions are marked with an asterisk in Table~I of the main paper.
Training and held-out functions can share a separability category while
differing in their base functions and other properties.
Decomposition results are precomputed to accelerate training.
Comparison methods follow the configurations in their original papers;
the NDA implementations use libraries such as
PyPop7~\cite{duan2024pypop7}.

\begin{table}[tbp]
\centering
\caption{Reported WCC-MMES training and evaluation settings.}
\label{supp:tab:settings}
\footnotesize
\begin{tabularx}{\linewidth}{@{}l X@{}}
\hline
Setting & Value \\
\hline
Training epochs & 90 \\
Training MaxFEs & $10^6$ per episode \\
Optimizer & Adam \\
Initial learning rate & $10^{-4}$; exponential decay \\
Discount factor $\gamma$ & 0.999 \\
Rollout length $n_{step}$ & 5 \\
PPO update epochs $K$ & 4 \\
Resource budgets & $\{5\!\times\!10^4,10^5,2\!\times\!10^5\}$ FEs \\
FeatureSE strength $\lambda$ & 0.2 \\
FeatureSE output biases & 2.0 \\
Fully/non-separable grouping & 20 subspaces \\
Main CEC2013LSGO tests & $1.5\!\times\!10^6$ FEs \\
DG2 decomposition allowance & $5\!\times\!10^5$ FEs, included in main budget \\
AOB tests & $1.5\!\times\!10^6$ FEs \\
Horizon tests & $3\!\times\!10^6$ FEs \\
NDA-transfer tests & $10^6$ FEs \\
Independent comparison runs & 30 per algorithm and problem \\
Significance test & Wilcoxon rank-sum; $\alpha=0.05$ \\
\hline
\end{tabularx}
\end{table}

The policy trained at $10^6$ FEs is reused for the generalization tests
without retraining. Numerical comparisons report the mean and standard
deviation over 30 independent runs.

\begin{table*}[tbp]
\centering
\footnotesize
\caption{WCC-MMES and comparison algorithms on CEC2013LSGO at $3\times10^6$ FEs.}
\label{supp:long-horizon}
\setlength{\tabcolsep}{2pt}
\resizebox{\textwidth}{!}{%
\begin{tabular}{r|c|ccc|ccccc|cc}
\hline
\multirow{2}{*}{Problem} & \shortstack{WCC-based\\Algorithms} & \multicolumn{3}{c|}{CC-based Algorithms} & \multicolumn{5}{c|}{NDAs} & \multicolumn{2}{c}{Hybrid Algorithms} \\
\cline{2-12}
 & WCC-MMES & DG2-CMAES & CC-CMAES & LCC-CMAES & MMES & sep-CMA-ES & LM-MA-ES & FCMAES & MetaES & HCC-ES & CC-SHADE-SL \\
\hline
\multicolumn{1}{c|}{1} & \cellcolor[rgb]{0.88,0.94,1.00}\textbf{2.03E-20} & 6.68E-20(+) & 7.38E+06(+) & 1.09E+05(+) & 1.18E+04(+) & 1.69E-17(+) & 1.63E+05(+) & 1.95E+05(+) & 9.09E+09(+) & 6.31E-20(+) & 2.58E-05(+) \\
 & \cellcolor[rgb]{0.88,0.94,1.00}{{$\pm$ 1.78E-20}} & {{$\pm$ 3.78E-21}} & {{$\pm$ 3.92E+06}} & {{$\pm$ 2.71E+04}} & {{$\pm$ 1.80E+03}} & {{$\pm$ 7.95E-19}} & {{$\pm$ 4.33E+04}} & {{$\pm$ 4.04E+04}} & {{$\pm$ 7.69E+08}} & {{$\pm$ 3.45E-21}} & {{$\pm$ 1.22E-04}} \\
\hline
\multicolumn{1}{c|}{2} & \cellcolor[rgb]{0.88,0.94,1.00}\textbf{1.35E+03} & 3.47E+03(+) & 2.13E+03(+) & 2.09E+03(+) & 1.26E+04(+) & 5.34E+03(+) & 5.98E+03(+) & 1.38E+04(+) & 4.05E+04(+) & 3.26E+03(+) & 1.98E+03(+) \\
 & \cellcolor[rgb]{0.88,0.94,1.00}{{$\pm$ 3.55E+02}} & {{$\pm$ 3.49E+02}} & {{$\pm$ 1.31E+02}} & {{$\pm$ 1.82E+02}} & {{$\pm$ 7.61E+02}} & {{$\pm$ 1.59E+02}} & {{$\pm$ 2.74E+02}} & {{$\pm$ 8.52E+02}} & {{$\pm$ 5.53E+03}} & {{$\pm$ 3.25E+02}} & {{$\pm$ 5.73E+02}} \\
\hline
\multicolumn{1}{c|}{3} & \cellcolor[rgb]{0.88,0.94,1.00}\textbf{2.00E+01} & 2.10E+01(+) & \textbf{2.00E+01($\approx$)} & 2.03E+01(+) & \textbf{2.00E+01($\approx$)} & 2.14E+01(+) & 2.15E+01(+) & 2.04E+01(+) & 2.16E+01(+) & 2.09E+01(+) & 2.08E+01(+) \\
 & \cellcolor[rgb]{0.88,0.94,1.00}{{$\pm$ 6.39E-03}} & {{$\pm$ 2.10E-02}} & {{$\pm$ 4.82E-03}} & {{$\pm$ 3.06E-02}} & {{$\pm$ 1.85E-03}} & {{$\pm$ 1.09E-01}} & {{$\pm$ 1.89E-01}} & {{$\pm$ 1.55E-01}} & {{$\pm$ 3.96E-01}} & {{$\pm$ 3.41E-02}} & {{$\pm$ 3.91E-02}} \\
\hline
\multicolumn{1}{c|}{4} & 1.32E+03 & \cellcolor[rgb]{0.88,0.94,1.00}\textbf{9.89E+02(-)} & 1.18E+11(+) & 2.27E+10(+) & 6.64E+07(+) & 3.73E+10(+) & 1.65E+11(+) & 4.78E+08(+) & 5.08E+12(+) & 9.81E+03(+) & 6.13E+09(+) \\
 & {{$\pm$ 8.74E+02}} & \cellcolor[rgb]{0.88,0.94,1.00}{{$\pm$ 5.73E+00}} & {{$\pm$ 6.05E+10}} & {{$\pm$ 1.33E+09}} & {{$\pm$ 5.45E+06}} & {{$\pm$ 8.85E+09}} & {{$\pm$ 3.41E+10}} & {{$\pm$ 4.98E+07}} & {{$\pm$ 2.43E+09}} & {{$\pm$ 3.46E+02}} & {{$\pm$ 4.68E+09}} \\
\hline
\multicolumn{1}{c|}{5} & \cellcolor[rgb]{0.88,0.94,1.00}\textbf{1.39E+06} & 1.58E+06(+) & 8.65E+06(+) & 1.02E+07(+) & 4.17E+06(+) & 1.64E+06(+) & 1.49E+06(+) & 3.82E+06(+) & 4.89E+07(+) & 1.60E+06(+) & 2.70E+06(+) \\
 & \cellcolor[rgb]{0.88,0.94,1.00}{{$\pm$ 1.21E+05}} & {{$\pm$ 1.62E+05}} & {{$\pm$ 2.68E+06}} & {{$\pm$ 1.01E+06}} & {{$\pm$ 7.67E+05}} & {{$\pm$ 1.52E+05}} & {{$\pm$ 4.58E+05}} & {{$\pm$ 8.33E+05}} & {{$\pm$ 8.21E+05}} & {{$\pm$ 1.32E+05}} & {{$\pm$ 7.29E+05}} \\
\hline
\multicolumn{1}{c|}{6} & \cellcolor[rgb]{0.88,0.94,1.00}\textbf{9.95E+05} & 1.06E+06(+) & 1.02E+06(+) & 1.05E+06(+) & \textbf{9.95E+05($\approx$)} & 1.06E+06(+) & 1.05E+06(+) & 9.96E+05(+) & 1.06E+06(+) & 1.06E+06(+) & 1.05E+06(+) \\
 & \cellcolor[rgb]{0.88,0.94,1.00}{{$\pm$ 4.08E-01}} & {{$\pm$ 1.22E+03}} & {{$\pm$ 1.56E+04}} & {{$\pm$ 2.08E+03}} & {{$\pm$ 2.14E-01}} & {{$\pm$ 1.78E+03}} & {{$\pm$ 1.35E+03}} & {{$\pm$ 9.38E+02}} & {{$\pm$ 1.63E+03}} & {{$\pm$ 9.98E+02}} & {{$\pm$ 2.25E+03}} \\
\hline
\multicolumn{1}{c|}{7} & \cellcolor[rgb]{0.88,0.94,1.00}\textbf{9.65E+01} & 1.38E+02(+) & 1.46E+09(+) & 4.91E+06(+) & 2.45E+04(+) & 2.36E+09(+) & 3.72E+06(+) & 1.65E+05(+) & 1.21E+15(+) & 9.01E+02(+) & 9.23E+05(+) \\
 & \cellcolor[rgb]{0.88,0.94,1.00}{{$\pm$ 4.97E+00}} & {{$\pm$ 2.20E+01}} & {{$\pm$ 2.15E+09}} & {{$\pm$ 3.11E+06}} & {{$\pm$ 1.58E+03}} & {{$\pm$ 1.87E+09}} & {{$\pm$ 4.13E+05}} & {{$\pm$ 3.53E+04}} & {{$\pm$ 7.78E+10}} & {{$\pm$ 2.44E+01}} & {{$\pm$ 8.83E+05}} \\
\hline
\multicolumn{1}{c|}{8} & 1.49E+07 & \cellcolor[rgb]{0.88,0.94,1.00}\textbf{2.42E+06(-)} & 4.53E+15(+) & 3.99E+14(+) & 1.84E+11(+) & 5.44E+14(+) & 1.71E+15(+) & 9.49E+11(+) & 1.46E+17(+) & 3.11E+06(-) & 1.21E+14(+) \\
 & {{$\pm$ 3.70E+06}} & \cellcolor[rgb]{0.88,0.94,1.00}{{$\pm$ 1.09E+06}} & {{$\pm$ 2.66E+15}} & {{$\pm$ 7.17E+13}} & {{$\pm$ 7.76E+09}} & {{$\pm$ 6.78E+14}} & {{$\pm$ 4.20E+14}} & {{$\pm$ 1.96E+11}} & {{$\pm$ 6.21E+15}} & {{$\pm$ 9.40E+05}} & {{$\pm$ 5.46E+13}} \\
\hline
\multicolumn{1}{c|}{9} & \cellcolor[rgb]{0.88,0.94,1.00}\textbf{1.13E+08} & 1.20E+08(+) & 7.60E+08(+) & 8.19E+08(+) & 4.41E+08(+) & 1.41E+08(+) & 2.20E+08(+) & 5.61E+08(+) & 3.73E+09(+) & 1.85E+08(+) & 2.27E+08(+) \\
 & \cellcolor[rgb]{0.88,0.94,1.00}{{$\pm$ 3.12E+07}} & {{$\pm$ 1.56E+07}} & {{$\pm$ 1.42E+08}} & {{$\pm$ 1.24E+08}} & {{$\pm$ 8.42E+07}} & {{$\pm$ 1.70E+07}} & {{$\pm$ 2.20E+07}} & {{$\pm$ 9.23E+07}} & {{$\pm$ 1.74E+06}} & {{$\pm$ 1.24E+07}} & {{$\pm$ 1.12E+08}} \\
\hline
\multicolumn{1}{c|}{10} & \cellcolor[rgb]{0.88,0.94,1.00}\textbf{9.05E+07} & 9.09E+07(+) & 9.41E+07(+) & 9.22E+07(+) & \textbf{9.05E+07($\approx$)} & 9.41E+07(+) & 9.40E+07(+) & 9.06E+07($\approx$) & 9.40E+07(+) & 9.10E+07(+) & 9.26E+07(+) \\
 & \cellcolor[rgb]{0.88,0.94,1.00}{{$\pm$ 6.43E+00}} & {{$\pm$ 2.77E+05}} & {{$\pm$ 5.04E+05}} & {{$\pm$ 2.02E+05}} & {{$\pm$ 3.47E+02}} & {{$\pm$ 1.66E+05}} & {{$\pm$ 2.77E+05}} & {{$\pm$ 3.98E+04}} & {{$\pm$ 6.90E+05}} & {{$\pm$ 2.87E+05}} & {{$\pm$ 3.44E+05}} \\
\hline
\multicolumn{1}{c|}{11} & \cellcolor[rgb]{0.88,0.94,1.00}\textbf{2.89E+04} & 2.94E+04(+) & 3.76E+11(+) & 4.38E+09(+) & 2.14E+06(+) & 3.85E+11(+) & 8.50E+07(+) & 1.41E+07(+) & 6.47E+16(+) & 3.34E+04(+) & 1.71E+08(+) \\
 & \cellcolor[rgb]{0.88,0.94,1.00}{{$\pm$ 2.54E+04}} & {{$\pm$ 4.36E+04}} & {{$\pm$ 2.85E+11}} & {{$\pm$ 4.83E+09}} & {{$\pm$ 1.87E+05}} & {{$\pm$ 1.36E+10}} & {{$\pm$ 1.46E+07}} & {{$\pm$ 1.96E+06}} & {{$\pm$ 3.23E+15}} & {{$\pm$ 3.96E+04}} & {{$\pm$ 9.20E+07}} \\
\hline
\multicolumn{1}{c|}{12} & 7.91E+02 & 1.18E+03(+) & 1.29E+03(+) & 1.31E+03(+) & 8.62E+02(+) & 8.66E+02(+) & 6.14E+02(-) & \cellcolor[rgb]{0.88,0.94,1.00}\textbf{5.34E+02(-)} & 1.04E+08(+) & 1.18E+03(+) & 1.32E+03(+) \\
 & {{$\pm$ 1.84E+01}} & {{$\pm$ 8.26E+01}} & {{$\pm$ 4.91E+03}} & {{$\pm$ 1.61E+02}} & {{$\pm$ 5.76E+02}} & {{$\pm$ 3.94E+01}} & {{$\pm$ 1.68E+02}} & \cellcolor[rgb]{0.88,0.94,1.00}{{$\pm$ 2.29E+02}} & {{$\pm$ 6.44E+08}} & {{$\pm$ 8.78E+01}} & {{$\pm$ 2.89E+02}} \\
\hline
\multicolumn{1}{c|}{13} & \cellcolor[rgb]{0.88,0.94,1.00}\textbf{2.46E+05} & 1.53E+08(+) & 5.79E+10(+) & 2.56E+07(+) & 2.89E+05(+) & 9.63E+10(+) & 8.74E+06(+) & 1.23E+06(+) & 9.44E+16(+) & 2.27E+06(+) & 9.75E+07(+) \\
 & \cellcolor[rgb]{0.88,0.94,1.00}{{$\pm$ 1.45E+04}} & {{$\pm$ 1.06E+08}} & {{$\pm$ 3.17E+10}} & {{$\pm$ 4.84E+06}} & {{$\pm$ 4.27E+04}} & {{$\pm$ 6.91E+10}} & {{$\pm$ 9.15E+05}} & {{$\pm$ 1.24E+05}} & {{$\pm$ 7.49E+15}} & {{$\pm$ 1.01E+06}} & {{$\pm$ 1.19E+08}} \\
\hline
\multicolumn{1}{c|}{14} & \cellcolor[rgb]{0.88,0.94,1.00}\textbf{5.81E+06} & 3.04E+07(+) & 8.53E+11(+) & 1.53E+08(+) & 7.91E+06(+) & 9.01E+09(+) & 1.02E+08(+) & 2.14E+07(+) & 9.47E+17(+) & 9.92E+06(+) & 9.41E+07(+) \\
 & \cellcolor[rgb]{0.88,0.94,1.00}{{$\pm$ 4.37E+06}} & {{$\pm$ 8.08E+06}} & {{$\pm$ 4.51E+11}} & {{$\pm$ 1.82E+07}} & {{$\pm$ 1.22E+06}} & {{$\pm$ 5.70E+09}} & {{$\pm$ 6.27E+06}} & {{$\pm$ 1.32E+06}} & {{$\pm$ 6.19E+15}} & {{$\pm$ 1.19E+06}} & {{$\pm$ 1.11E+08}} \\
\hline
\multicolumn{1}{c|}{15} & \cellcolor[rgb]{0.88,0.94,1.00}\textbf{3.89E+05} & 3.54E+13(+) & 3.06E+08(+) & 1.85E+07(+) & 1.29E+06(+) & 2.66E+08(+) & 9.88E+06(+) & 6.61E+06(+) & 2.35E+15(+) & 2.79E+13(+) & 2.00E+06(+) \\
 & \cellcolor[rgb]{0.88,0.94,1.00}{{$\pm$ 1.67E+05}} & {{$\pm$ 6.35E+12}} & {{$\pm$ 3.06E+08}} & {{$\pm$ 3.59E+06}} & {{$\pm$ 1.35E+05}} & {{$\pm$ 5.27E+07}} & {{$\pm$ 2.35E+06}} & {{$\pm$ 9.06E+05}} & {{$\pm$ 1.62E+12}} & {{$\pm$ 1.61E+13}} & {{$\pm$ 5.73E+05}} \\
\hline
\multicolumn{1}{c|}{+/$\approx$/-} & NA & 13/0/2 & 14/1/0 & 15/0/0 & 12/3/0 & 15/0/0 & 14/0/1 & 13/1/1 & 15/0/0 & 14/0/1 & 15/0/0 \\
\hline
\end{tabular}
}
\par\smallskip
\begin{minipage}{\textwidth}\footnotesize
Entries report means and standard deviations over 30 runs. Symbols $+$, $\approx$, and $-$ indicate WCC-MMES is significantly better, comparable, and worse, respectively (Wilcoxon rank-sum, $\alpha=0.05$). The final row reports win/tie/loss counts.
\end{minipage}
\end{table*}

\subsection{Ablation Study}
\label{supp:ablation}
To further examine where the performance of WCC-MMES comes from, we conduct a comprehensive ablation study from four aspects: action space, state space, reward function, and network architecture.

\subsubsection{Experimental Protocol}
For the action space ablation, we consider both random and fixed variants. In the random variants, either both action components, only the mode (Random($a_{\text{mode}}$)), or only the resource (Random($a_{\text{res}}$)) are randomized. In the fixed variants, either the mode is fixed to CC or NDA while the resource is controlled by the agent (Fixed($a_{\text{mode}}$)), or the resource budget is fixed while the mode is controlled by the agent (Fixed($a_{\text{res}}$)). For the state space ablation, we remove each major state group individually, including problem features (W/O $S_{\text{pro}}$), convergence features (W/O $S_{\text{conv}}$), resource features (W/O $S_{\text{res}}$), and mode history features (W/O $S_{\text{mode}}$). For the reward ablation, the proposed log-ratio reward is replaced by Difference $r_t= cost^*_{t-1}-cost^*_{t}$, Sparse Sign $$r_{t}=\left\{\begin{array}{ll}
                            1, & cost_{t-1}^{*}-cost_{t}^{*}>\epsilon \\                             0, & \text { otherwise }
                             \end{array} \quad \epsilon=1\text{E}-2\right.$$, and Linear $r_{t}=\frac{cost_{t-1}^{*}-cost_{t}^{*}}{\left|cost_{t-1}^{*}\right|}$ rewards. For the network ablation, we remove FeatureSE (W/O FeatureSE), replace the auto-regressive policy with a simple MLP, or remove both. For fair comparison, the final costs are min–max normalized within each category, and the mean normalized performance over all problems is reported. All ablated variants are trained and evaluated under the same protocol as Section~III-B1 of the main paper.

\subsubsection{Normalization and reporting}

For category $c$, let $\mathcal{A}_c$ contain WCC-MMES and the variants
compared in that category, and let $\bar f_{p,a}$ denote the mean final
cost of variant $a$ on problem $p$. The per-problem extremes are
$f^{\min}_{p,c}=\min_{a\in\mathcal{A}_c}\bar f_{p,a}$ and
$f^{\max}_{p,c}=\max_{a\in\mathcal{A}_c}\bar f_{p,a}$.
The inverse min--max score is
\begin{equation}
\label{supp:ablation-normalization}
q_{p,a}^{(c)}=
\begin{cases}
1-\dfrac{\bar f_{p,a}-f^{\min}_{p,c}}
{f^{\max}_{p,c}-f^{\min}_{p,c}},&f^{\max}_{p,c}>f^{\min}_{p,c},\\
1,&f^{\max}_{p,c}=f^{\min}_{p,c}.
\end{cases}
\end{equation}
For the problem set $\mathcal{P}_c$ evaluated in category $c$, the
reported mean normalized performance is
\begin{equation}
\label{supp:ablation-mean-score}
Q_a^{(c)}=\frac{1}{|\mathcal{P}_c|}
\sum_{p\in\mathcal{P}_c}q_{p,a}^{(c)}.
\end{equation}
Costs are normalized separately for each problem and ablation category
before averaging. Higher scores indicate better performance within
the corresponding category.

\begin{figure}[tbp]
\centering
\includegraphics[width=\linewidth]{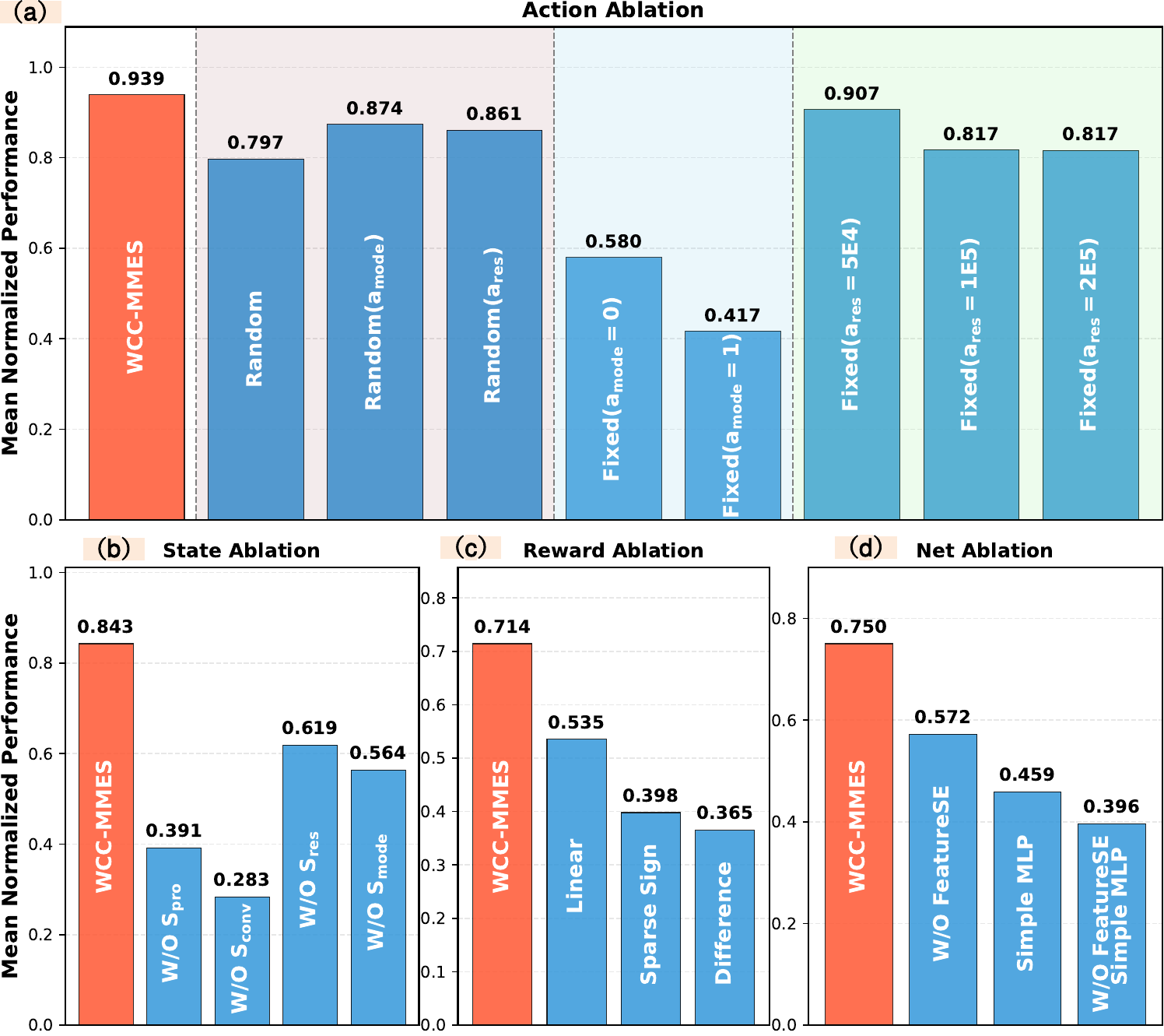}
\caption{Ablation study, including (a) action space ablation, (b) state space ablation, (c) reward function ablation, and (d) network architecture ablation.}
\label{supp:fig:ablation}
\end{figure}

\subsubsection{Ablation Analysis}
The results are shown in Figure~\ref{supp:fig:ablation}. 
a) Action space. 
The key takeaway from Figure~\ref{supp:fig:ablation}(a) is that WCC benefits from jointly learning both the optimization mode and the resource allocation. Fixing the mode leads to the worst performance, indicating that neither CC nor NDA is universally optimal across all problems and stages. Randomizing either the mode selector or the resource allocator also degrades performance, showing that the learned decisions cannot be replaced by arbitrary switching. 

b) State space.
The key takeaway from Figure~\ref{supp:fig:ablation}(b) is that the four state groups provide complementary information rather than redundant signals. Removing any major state group leads to a clear performance drop: problem features provide structural priors such as dimensionality and decomposability, convergence features capture the current optimization dynamics and cause the largest degradation when removed, while resource and mode-history features describe budget consumption and historical mode preference, improving the timing and stability of the learned switching policy.

c) Reward function. 
The key takeaway from Figure~\ref{supp:fig:ablation}(c) is that a stable and scale-aware reward is critical for learning across heterogeneous LSGO problems. Replacing the proposed log-ratio reward with Difference, Sparse Sign, or Linear rewards consistently worsens performance: Difference rewards are sensitive to objective scales, Sparse Sign rewards discard the magnitude of improvement and depend on threshold selection, and Linear rewards may still suffer from numerical instability. 

d) Network architecture. 
The key takeaway from Figure~\ref{supp:fig:ablation}(d) is that both FeatureSE and the auto-regressive policy contribute to WCC's performance. Removing FeatureSE degrades performance, confirming the usefulness of adaptively reweighting heterogeneous state features. Replacing the auto-regressive policy with a simple MLP also hurts performance, because WCC naturally involves a hierarchical decision process: the agent should first select CC or NDA and then allocate FEs conditioned on the selected mode. The largest degradation occurs when both FeatureSE and the auto-regressive structure are removed, indicating that heterogeneous state encoding and mode-conditioned resource allocation jointly support effective CC deployment.

\bibliographystyle{IEEEtran}
\bibliography{ref}